%% file: main.tex
\documentclass[10pt,twocolumn,letterpaper]{article}

\usepackage{iccv}              % To produce the CAMERA-READY version
\input{preamble}

\definecolor{iccvblue}{rgb}{0.21,0.49,0.74}
\usepackage[pagebackref,breaklinks,colorlinks,allcolors=iccvblue]{hyperref}

\def\paperID{6957} % *** Enter the Paper ID here
\def\confName{ICCV}
\def\confYear{2025}

\title{Weakly-Supervised Learning of Dense Functional Correspondences}

\author{Stefan Stojanov* \quad 
Linan Zhao* \quad
Yunzhi Zhang \quad
Daniel L. K. Yamins \quad
Jiajun Wu \\
Stanford University
}

\begin{document}
\maketitle
\def\thefootnote{*}\footnotetext{Equal contribution.}\def\thefootnote{\arabic{footnote}}

\input{sec/0_abstract}
\input{sec/1_intro}
\input{sec/2_related_work}

\input{sec/3_problem_formulation}

\input{sec/4_approach}
\input{sec/5_experiments}

\input{sec/7_conclusion}

\clearpage
\paragraph{Acknowledgments.} This work is in part supported by NSF RI \#2211258 and \#2338203, NSF CAREER \#1844724, NSF NCS-FR \#2123963, ONR YIP N00014-24-1-2117, ONR MURI N00014-22-1-2740, ONR N00014-20-1-2589, ONR MURI N00014-21-1-2801, ONR MURI N00014-24-1-2748, and Simons Foundation grant SFI-AN-NC-GB-Culmination-00002986-05. We also thank the Stanford HAI for their support with computing resources.

{
    \small
    \bibliographystyle{ieeenat_fullname}
    \bibliography{main}
}

%\appendix

\section*{Appendix}
\input{sec/8_supp_arxiv}

\end{document}

%% file: preamble.tex
\usepackage{comment}
\usepackage{multirow}
\usepackage{makecell}
\usepackage{bbm}
\usepackage{tabularx}
\usepackage{xurl}

\usepackage{framed}
 {\endMakeFramed}
\definecolor{shadecolor}{gray}{0.75}

\renewcommand{\paragraph}[1]{\vspace{0.1cm}\noindent\textbf{#1}}

%% file: sec/0_abstract.tex
\begin{abstract}
Establishing dense correspondences across image pairs is essential for tasks such as shape reconstruction and robot manipulation. In the challenging setting of matching across different categories, the function of an object, \ie, the effect that an object can cause on other objects, can guide how correspondences should be established. This is because object parts that enable specific functions often share similarities in shape and appearance. We derive the definition of \emph{dense functional correspondence} based on this observation and propose a weakly-supervised learning paradigm to tackle the prediction task. The main insight behind our approach is that we can leverage vision-language models to pseudo-label multi-view images to obtain functional parts. We then integrate this with dense contrastive learning from pixel correspondences to distill both functional and spatial knowledge into a new model that can establish dense functional correspondence. Further, we curate synthetic and real evaluation datasets as task benchmarks. Our results demonstrate the advantages of our approach over baseline solutions consisting of off-the-shelf self-supervised image representations and grounded vision language models.\footnote{Accepted at ICCV 2025. Project website: \url{https://dense-functional-correspondence.github.io/}}
\end{abstract}

%% file: sec/1_intro.tex
\section{Introduction}
\label{sec:intro}

Finding pixel correspondence across image pairs is fundamental for object understanding and is critical for applications like shape reconstruction~\cite{lindenberger2021pixel, truong2023sparf, mai2024tracknerf,lao2023corresnerf}, editing~\cite{gu2024videoswap}, and object manipulation in robotics~\cite{ju2024robo, dipalo2024dinobot, florence2018dense,jiang2024doduo}.
This task requires reasoning beyond visual similarity in local appearance, geometry, and texture across images. It also involves structural similarity, \eg, the part-whole relationships of objects and their part components, and semantic similarity, \eg, the functional properties of parts of objects. 

These aspects of similarity are essential for learning efficient generalizable systems for downstream applications. For example, in imitation learning in robotics, human demonstrations are a scarce and valuable data source. Given a demonstration with an object, such as pouring with a kettle, establishing dense functional correspondence with another object that supports this function, \eg, a bottle, enables the efficient transfer of the demonstration.

It becomes harder to find dense correspondence when the input images shift from being two views of the same object to different objects from the same category, and finally to objects from distinct categories, as the visual similarity becomes less apparent.
This work focuses on the most challenging scenario with objects from different categories.
We aim to establish dense pixel-level correspondence between pairs of images containing objects with parts whose shape enables the execution of similar functions. Specifically, by ``function'', we refer to the effect one object can have on another object or substance, \eg, the function ``cut-with'' for a knife and a spatula or ``hang-onto'' for objects with hooks.

Practically, training and evaluation for this task are challenging due to the lack of labeled data. 
Supervised training at scale is infeasible because manual dense correspondence labeling is intractable, emphasizing the need for a self- or weakly-supervised approach. 
For evaluation, while datasets exist for dense within-category correspondence~\cite{taniai2016joint, wang2019normalized, kwrishnan2024omninocs} and sparse functional keypoint correspondence across categories~\cite{lai2021functional}, there is still no established task or dataset for dense correspondence across categories. In this work, we make progress toward addressing both the challenges of training and evaluation.

\input{figures_tex/intro_figure}

The key insight behind our training approach is that the capabilities of self-supervised image representations like DINOv2~\cite{oquab2023dinov2} or Stable Diffusion~\cite{rombach2022high} and vision language models (VLMs)~\cite{wang2023cogvlm, huang2024manipvqa} are complementary but individually insufficient for solving this task. On the one hand, surprisingly accurate dense correspondences can be established using image features from pre-trained self-supervised models. This works well when the input images contain visually similar object instances from the same category, \eg, two cats or two cars~\cite{zhang2023talefeaturesstablediffusion}. However, the accuracy decreases for the more generic scenario when objects come from distinct categories. 
On the other hand, VLMs can detect the bounding boxes of object parts with similar functions in a zero-shot manner~\cite{wang2023cogvlm, huang2024manipvqa} but cannot perform fine-grained reasoning about correspondences across objects. 

We distill the strengths of each approach into a new model using a scalable technique that requires minimal human supervision. Specifically, we first obtain multi-view-consistent pseudo labels of functionally relevant regions of 3D object assets~\cite{deitke2023objaverse} using an off-the-shelf grounded VLM~\cite{wang2023cogvlm}. We then combine these labels with multi-view correspondences~\cite{stojanov2022learning, florence2018dense} in a contrastive learning framework building on pre-trained DINOv2~\cite{oquab2023dinov2} feature extractor. For evaluation, we define the dense functional 2D correspondence task and develop an annotation procedure based on aligning 3D object pairs in functionally equivalent poses.

In sum, we define the task of dense functional correspondence as a means for investigating cross-category dense correspondence. We then curate synthetic and real-world evaluation datasets for this task. We further propose a scalable, weakly-supervised method leveraging vision foundation models, which empirically outperforms baselines.

%% file: figures_tex/intro_figure.tex
\begin{figure*}[t]
    \centering
    \includegraphics[width=\linewidth]{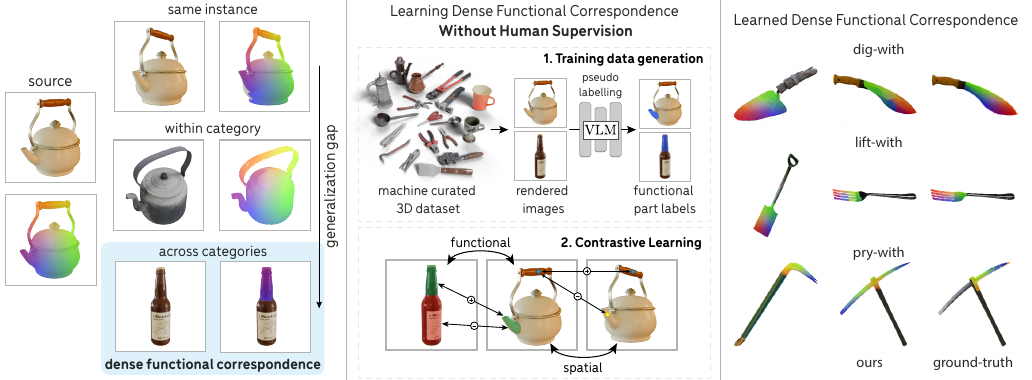}
    \vspace{-1.5em}
    \caption{
    \textbf{Dense Functional Correspondence} refers to establishing dense correspondences across object instances based on function similarity (\eg, ``pour-with''). This task is especially challenging when objects have visually different but functionally similar parts, requiring both semantic understanding, \ie, identifying which parts can perform the same function, and structural understanding, \ie, establishing dense correspondence across the parts at a surface-point-level based on functionally equivalent alignment. We propose a method to learn such correspondences with little human supervision, leveraging automated data curation and annotation, and dense contrastive learning.}
    \label{fig:intro_figure}
    \vspace{-1.75em}
\end{figure*}

%% file: sec/2_related_work.tex
\section{Related Work}
\label{sec:lit-review}
Learning representations to establish dense functional correspondence requires fine-grained structural and semantic visual reasoning about objects. 
The most relevant prior works come from the object-level correspondences and affordance learning research domains. 
We also review recent work on vision foundation models and VLMs, focusing on works relating to fine-grained object understanding.

\paragraph{Learning Correspondences.}
For this work, it is relevant to categorize correspondence learning methods based on their degrees of generalization. 
For generalizing across geometric scene transforms, works on multi-view correspondences aim to match different views of the same scene~\cite{sarlin2020superglue, sun2021loftr, jiang2021cotr}, whereas optical flow techniques match consecutive video frames~\cite{ilg2017flownet, teed2020raft, huang2022flowformer}. For generalizing within categories, NOCS-style representations~\cite{wang2019normalized, kwrishnan2024omninocs, wan2023socs} enable dense matching across instances of a category, whereas learning sparse keypoints~\cite{min2019spair, sun2023misc210k} enables sparse matching based on pre-defined semantic keypoint taxonomies.
For generalizing across categories, Lai et al.~\cite{lai2021functional} propose matching based on object function by learning five keypoints per object function category. The main drawback of keypoint-based correspondences is the requirement for a keypoint taxonomy, which by definition limits such techniques' capability to capture nuanced similarities across highly dissimilar objects (\eg, a bottle and a kettle). Through our dense functional correspondence formulation, we overcome the limitation of keypoint definitions and enable higher precision in downstream applications.

\paragraph{Learning Affordances.} 
In his seminal work~\cite{gibson1979ecological}, James J. Gibson defines affordances as objects' ``opportunities for interaction.'' Various object affordance formalisms have been developed in computer vision and robotics, such as estimating grasps~\cite{brahmbhatt2019contactdb, murali2021same, fang2020graspnet, mousavian20196}, and localizing affordance regions in 2D~\cite{myers2015affordance, chuang2018affordancesfromimages, do2018affordancenet,  luo2022agd20k, nagarajan2019grounded, nguyen2017affordance} and 3D~\cite{deng20213d, yang2023grounding, guo2023handal} through bounding boxes and segments~\cite{do2018affordancenet, myers2015affordance,nguyen2017affordance,nagarajan2019grounded}, heatmaps~\cite{myers2015affordance, luo2022agd20k, deng20213d} or keypoints~\cite{turpin2021gift, qin2020keto, xu2021affordance}. 
Early works adopt a fully supervised learning paradigm~\cite{do2018affordancenet, myers2015affordance, brahmbhatt2019contactdb}, while more recent works aim to use less supervision by learning from human object interaction videos~\cite{nagarajan2019grounded}, egocentric videos~\cite{li2024learning} or unlabeled exocentric images~\cite{li2023locate, luo2022agd20k}. 
Our work has two key distinctions: First, affordance heatmaps or segments identify object regions or parts in individual images. They do not allow for fine-grained spatial correspondence across object parts in different images (\eg, can identify the blades of two knives but cannot find correspondences for pixels between the tips or edges of the blades). 
Second, our focus is on object function -- the effect an object can cause on something else, rather than the broader concept of affordance, which emphasizes potential interactions with a specific object instance (\eg, striking with a hammer vs. holding). Last, our goal is to learn dense functional correspondence in a weakly-supervised manner, without relying on human annotations of ground-truth correspondences.

\paragraph{Vision Foundation Models.}
Recent developments in large-scale language~\cite{touvron2023llama, 2020t5, devlin2018bert} and image~\cite{radford2021learning, zhai2023sigmoid} pre-training have led to the development of vision-language models (VLMs) capable of strong zero-shot performance through vision-question answering~\cite{lin2023sphinx, wang2023cogvlm}, which have been adapted to reasoning about functional affordances and grasping in robotics~\cite{ding2024preafford, huang2024manipvqa, qian2024affordancellm, yuan2024robopoint}. 
Powerful correspondence representations have been found to emerge~\cite{amir2021deep, tang2023emergent} in DINO~\cite{oquab2023dinov2, caron2021emerging} and Stable Diffusion~\cite{rombach2022high}, which have led to direct applications in low-shot affordances~\cite{ju2024robo} and object manipulation~\cite{ju2024robo, kuang2024ram, dipalo2024dinobot, papagiannis2024miles, di2024effectiveness}.  
In this work, we leverage the complementary characteristics of VLMs and self-supervised image models to go beyond their individual capabilities for dense functional correspondence.

%% file: sec/3_problem_formulation.tex
\section{Dense Functional Correspondence}
\label{sec:problem-definition}

Distinct object categories with similar functionality, \eg, a ``kettle'' and a ``bottle'' which can both pour liquid, may have different visual shapes and appearances as well as distinct part organizations. However, individual parts that serve the specific functionality of interest, \eg, the spout of a kettle and the mouth of a bottle in this example, have a higher resemblance with each other than at the overall object level. 
Such consistency is a consequence of how form follows function -- object parts that fulfill a specific function tend to remain consistent across objects, even if other parts vary greatly in shapes and arrangements. The part-level consistency provides a crucial ground from which we can derive the definition of functional correspondence (\Cref{ssec:definition}) and develop a corresponding evaluation data curation pipeline to benchmark this task (\Cref{sec:get-gt}).

\input{figures_tex/annotation_figure}

\input{figures_tex/pseudo_labeling}

\subsection{Problem Definition}
\label{ssec:definition}

We refer to the effect that an object causes on other objects or substances as an ``object function.'' This concept has been widely studied in model generalization in visual computing~\cite{thompson2021shape, zhu2015understanding, lai2021functional, laga2013geometry} and the development of categorization in humans~\cite{landau1998object, kemler2000two, ware2010form}. Examples are shown in \Cref{fig:intro_figure}, \eg, ``pour with.'' When executing a function with an object, such as pouring with a kettle, the functional part (the spout) follows a specific 3D trajectory. To replicate this function with a different object, \eg, a bottle, the neck of the bottle would be aligned with the spout and follow the same trajectory.
This illustrates how the \emph{same} object function is fulfilled with \emph{different} objects via aligning functionally equivalent parts, which is central to robotic applications with imitation learning approaches~\cite{zhu2024vision, heppert2024ditto, dipalo2024dinobot}.

The above observations lead us to define dense functional correspondence through \emph{3D object alignment based on functionally equivalent parts.}
Specifically, given two objects (\eg, a kettle and a bottle) and an object function (\eg, ``pour with''), the objects are aligned if and only if the parts that fulfill this function (\eg, the kettle spout and the bottle neck) are spatially close to each other.
The alignment induces an image-space distance: for any pair of pixels on the functional parts of two objects, the pixels are in \emph{functional correspondence} if their respective surface points are close in 3D when the objects are aligned.
Since this distance is defined at the pixel level, it is inherently \emph{dense}.

Formally, the input consists of an object function $\mathcal{F}$ and an image pair $(I_1, I_2)$, where each image is a view of a 3D object $O_1$ and $O_2$. Let $\pi^{-1}: I\rightarrow O$ represent the back-projection function that maps an image pixel to the corresponding 3D object surface point. We define $M(O;\mathcal{F})$ as the functional part of object $O$ responsible for executing $\mathcal{F}$, and let $M(I;\mathcal{F})$ be its projected 2D mask in the image. In our setup, the functional parts of both objects, $M(O_1;\mathcal{F})$ and $ M(O_2; \mathcal{F})$, are assumed to be aligned in 3D such that they follow the same trajectory when performing $\mathcal{F}$. We therefore define \emph{dense functional correspondence} as a mapping $f(I_1, I_2; \mathcal{F}): M(I_1; \mathcal{F})\rightarrow M(I_2;\mathcal{F})$ that minimizes $\sum_{p \in   M(I_1; \mathcal{F})} ||\pi^{-1}(p) - \pi^{-1}(f(p))||_2$. This ensures that pixel pairs in functional correspondence are from spatially close locations in 3D when the objects are aligned.

\subsection{Evaluation Dataset Curation}
\label{sec:get-gt}
The problem definition in \Cref{ssec:definition} provides a guiding principle to obtain ground truth annotations for dense functional correspondence in image pairs by \emph{aligning objects in 3D}. We introduce the annotation procedure and use it to construct both synthetic and real-world evaluation datasets for quantitative evaluation in \Cref{sec:experiments}.

\paragraph{Annotation Procedure.}
To obtain ground-truth functional correspondence for an image pair, we assume each object is rendered from a known 3D asset. By aligning the two assets in 3D, we derive dense pixel correspondences between the images. This procedure eliminates the need for manual dense 2D labeling, enabling large-scale evaluation. An overview is shown in \Cref{fig:annotation_figure}.

Specifically, given two 3D meshes of objects supporting the same function, we first align them based on their functional parts and annotate a 3D bounding box around each functional part. Then, for a pair of rendered images, we unproject pixels from the functional parts onto the object surfaces and compute 3D distances between these points to perform minimum-cost matching. Pixels corresponding to visible surface points that are spatially close in 3D are matched. A detailed description of the annotation procedure is provided in the Supplement.

\paragraph{Synthetic Evaluation Dataset.}
We use the 3D assets from Objaverse~\cite{deitke2023objaverse}, a large collection of diverse 3D models, to obtain a synthetic evaluation dataset. 
We hand-label 3D annotations for 950 pairs of assets from Objaverse spanning 24 functions, selected for quality and diversity. See \Cref{sec:pseudo-label} for how the assets and functions are selected. As such, 85\% of the ground-truth pairs contain across-category correspondences. From these 3D annotations, we derive 1,800+ unique 2D image pairs rendered from the 3D assets, with ground truth dense functional correspondences. 

\paragraph{Real Evaluation Dataset.}
Setting up a real-world benchmark is crucial for measuring model performance on real images. Thus, we utilize the HANDAL dataset~\cite{guo2023handal}, which contains images and 3D reconstructions of real-world objects. After manually fixing the geometry of the 3D scanned assets, \eg, the missing concavities of mugs, pots and pans, we label 190 asset pairs spanning 13 functions. This results in a real evaluation dataset of 500+ unique 2D real image pairs with ground-truth functional correspondence labels.

%% file: figures_tex/annotation_figure.tex
\begin{figure}[t]
    \centering
    \includegraphics[width=\linewidth]{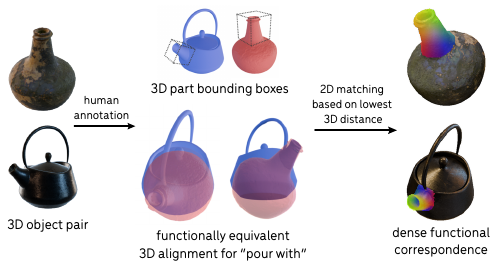}
    \vspace{-21pt}
    \caption{\textbf{Annotation Pipeline (Evaluation Only).} Given a 3D object pair (left) and a function (``pour-with''), we annotate the functional alignment of two objects by aligning the functional parts in 3D (middle). Afterward, we derive dense 2D correspondences (right) based on 3D distances of corresponding object surface points, with matching pixels shown in the same color.}
    \vspace{-1.7em}
    \label{fig:annotation_figure}
\end{figure}

%% file: figures_tex/pseudo_labeling.tex
\begin{figure*}[t]
    \centering
    \includegraphics[width=\linewidth]{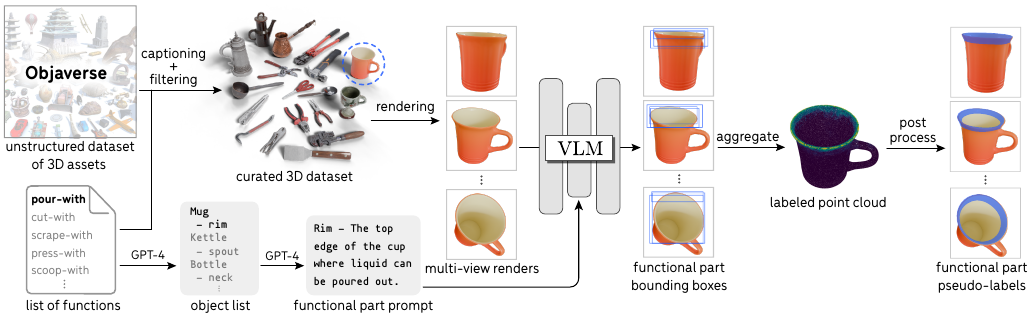}
    \vspace{-2.1em}
    \caption{\textbf{Training Data Curation via VLM Pseudo Labeling.} 
    Given a large unstructured dataset like Objaverse~\cite{deitke2023objaverse}, we leverage off-the-shelf VLMs to curate and label the functional parts. Specifically, GPT-4~\cite{gpt4} generates category-specific functional part prompts, and CogVLM~\cite{wang2023cogvlm} produces bounding box proposals for multi-view image renderings, which are aggregated onto a 3D point cloud. The point cloud is post-processed to produce pixel-level functional part labels for training.}
    \label{fig:labeling_pipeline}
    \vspace{-15pt}
    % \vspace{-1.5em}
\end{figure*}

%% file: sec/4_approach.tex
\section{Approach}
\label{sec:approach}

Our goal is to develop a scalable learning framework for dense functional correspondences without relying on human-labeled ground truth. Since this task requires both semantic and structural knowledge, we distill from off-the-shelf VLMs to obtain pseudo-labeled training data (\cref{sec:pseudo-label}), which is further combined with dense spatial correspondences from synthetic data in a contrastive learning framework (\cref{sec:contrastive}). This approach enables the model to generalize to real-world data, as we will show in \cref{sec:experiments}.

\subsection{Dataset of Pseudo-labeled Functional Parts}
\label{sec:pseudo-label}

A dataset for learning dense functional correspondences at scale requires a diverse source of object images, a diverse taxonomy of functions and associated functional parts, and a low-cost, reliable means for part labeling. 

\paragraph{Image Data.} Our approach requires a large and diverse multi-view image dataset where functional parts are visible. Existing multi-view object datasets~\cite{wu2023omniobject3d, reizenstein2021common, yu2023mvimgnet, zhu2023egoobjects} are suboptimal because they have few desired objects like tools and utensils, the objects are in canonical poses that may not reveal functional parts, or are placed in cluttered contexts where occlusions often occur. To overcome this, we render high-quality images from the Objaverse~\cite{deitke2023objaverse} dataset using ray-tracing and HDRI environments~\cite{hdrihaven} in Blender~\cite{blender}, obtaining arbitrary amounts of diverse multi-view data. 

\paragraph{Object and Function Taxonomy.} To curate relevant object assets for our training dataset, we prompt GPT-4~\cite{gpt4} for common functions and refer to object functions studied in~\cite{murali2021same, lai2021functional}. Then, we prompt GPT-4 to generate a comprehensive list of object categories for each function. 
After deduplication and manual filtering, our taxonomy has 24 functions and 160 object categories. 

\paragraph{Object Asset Selection.} To retrieve relevant assets from the noisy-labeled Objaverse dataset based on the list of object categories, we utilize asset captions from Caption3D~\cite{luo2024caption3d}. We use Llama 3.1~\cite{dubey2024llama} to summarize the captions into category names and use Llama word embeddings to match the summaries to our category list. Finally, we prompt Llama to verify these matches. To ensure diversity, we cap each category at 200 assets. To ensure quality, we manually filter the retrieved assets to obtain 8,285 assets in total, 80\% of which are used for training. Details about prompting, filtering, and the taxonomy are included in the Supplement.

\paragraph{Functional Part Pseudo-Labeling.} Labeling data at scale using large pre-trained models has been shown as an effective approach for achieving high performance with minimal human effort~\cite{yang2024depth, vemulapalliknowledge}. The key elements for success are a sufficiently accurate pre-trained model and a low-cost and reliable procedure for rejecting low-quality labels. Grounded VLMs~\cite{wang2023cogvlm, lin2023sphinx, deitke2024molmo} have shown remarkable capabilities for zero-shot prompt-based object detection. We, therefore, use the 17B grounded CogVLM~\cite{wang2023cogvlm} model, which has state-of-the-art referring expression detection performance. For an overview of the pseudo-labeling pipeline, see~\Cref{fig:labeling_pipeline}. Given our list of object categories and functions, we prompt GPT-4 to obtain the names and appearance descriptions of functional parts to serve as prompts for CogVLM, which we then manually filtered and deduplicated. Because functional part names can be different across categories (\eg, the spout of a kettle vs. the neck of a bottle), we generate these functional part lists separately for each category. We empirically found that prompting CogVLM with part names and appearance descriptions significantly improves the bounding box predictions. 

Given a set of rendered views for an object and a functional part text prompt, we generate bounding box predictions with CogVLM~\cite{wang2023cogvlm}, which vary due to sampling in VLM inference. The accuracy of the bounding boxes also depends on viewpoint because of part pose and visibility. To aggregate these possibly noisy labels and obtain a final part label, we sample a dense point cloud on the surface of the object, and accumulate the 2D labels across views onto the 3D points. We post-process these labeled point clouds to generate 2D masks for views rendered for training.

This dataset curation and pseudo-labeling procedure allows us to generate a large dataset of functional part segmentation labels with relatively little human effort, which was mostly necessary for prompt engineering and quality control. In this work, we apply this approach on the $\approx 600$K labeled meshes from Caption3D, but it is straightforward to scale up to the millions of meshes in Objaverse-XL~\cite{deitke2024objaverse}.

\subsection{Learning Dense Functional Correspondence}
\label{sec:contrastive}

To learn dense functional correspondence, we train a feature embedding that captures both the high-level function semantics and the structural similarity between functional parts. For instance, given a bottle and a kettle for the function ``pour-with," the features for the neck of the bottle and the spout of the kettle should be similar. Moreover, the mouth of the bottle and the tip of the kettle spout should be in correspondence, as well as the bottom of the bottle's neck and the bottom of the kettle's spout. To achieve this, we train a function-conditioned network on top of frozen DINOv2~\cite{oquab2023dinov2} and CLIP~\cite{radford2021learning} (illustrated in~\Cref{fig:architecture_diagram}), that is applied at the local feature level. Because of significant developments in object segmentation~\cite{kirillov2023segment, ke2024segment} and our focus on object-level understanding, we assume that the input images consist of segmented objects.

\paragraph{Function-Conditioned MLP.} Given an image and a function, we first extract the image features from the last three blocks of DINOv2 and the function conditioning from CLIP text embeddings. We average the DINOv2 features from each block using learned weights into a single feature grid, and use bilinear interpolation to obtain a feature vector for each pixel location. Then, we concatenate the image feature with the CLIP embedding of the function and pass it through a 3-layer MLP, which produces the final feature at each pixel location. This network can be thought of as a function-conditioned version of the final projection layer used in contrastive learning~\cite{chen2020simple, he2020momentum}. We parameterize our model as $g_{\theta}(p|I, \mathcal{F})$, which outputs the normalized feature of pixel $p$ on image $I$ conditioned on the function $\mathcal{F}$.

\input{figures_tex/architecture_diagram}

We also investigate the option of adding an extra fully connected layer that maps the output feature vector to a prediction for the functional part mask. This allows us to obtain a binary functional part mask at inference time. 

\paragraph{Functional Part Contrastive Learning.} To distill the knowledge of functional part semantics from the VLM, we use contrastive learning based on the pseudo-labeled functional part masks. The parts from two objects that can be used to perform the same function should share a more similar embedding space. Specifically, given two images, $I_1$ and $I_2$ of objects that can perform the same function $\mathcal{F}$, let the functional part segments be $P_1^{+}$ and $P_2^{+}$. Then, define the rest of the objects' pixels as $P_1^{-}$ and $P_2^{-}$. Learning correspondence requires the pixels in $P_1^{+}$ to be similar to the ones in $P_2^{+}$ but different from the ones in $P_2^{-}$. In addition, to encourage the model to focus on the functionally relevant regions of objects, we add a term that pushes the features of $P_1^{-}$ away from that of $P_2^{-}$.

Let $\text{sim}(x,y | I_1, I_2, \mathcal{F}) = g(x|I_1,\mathcal{F}) \cdot g(y | I_2, \mathcal{F})$ represent the feature similarity between pixel $x$ on image $I_1$ and pixel $y$ on image $I_2$ when conditioned on function $\mathcal{F}$. For brevity, we short-hand it to $\text{sim}(x,y)$ below. The infoNCE loss~\cite{oord2019representationlearningcontrastivepredictive} for the function-part contrastive learning given $(p_1^{+}, p_1^{-}, p_2^{+}, p_2^{-}) \in (P_1^{+}, P_1^{-}, P_2^{+}, P_2^{-})$ is thus 
\begin{equation}
    \mathcal{L}_{\text{func}} = -\log \frac{e^{\text{sim}(p_1^{+}, p_2^{+}) / \tau}}{e^{\text{sim}(p_1^{+}, p_2^{+}) / \tau} + e^{\text{sim}(p_1^{+}, p_2^{-}) / \tau} + e^{\text{sim}(p_1^{-}, p_2^{-}) / \tau}}
\end{equation}
for temperature $\tau$.

When the model predicts functional part masks, we add a binary cross-entropy loss $\mathcal{L}_{\text{mask}}$ to compare the predicted mask with the pseudo-labeled functional part segment.

\paragraph{Part Structure via Multi-view Contrastive Learning.} If we train the embedding with only the functional part contrastive loss, we inevitably run into mode collapse issues. That is, the whole spout of the kettle would have the same features regardless of the pixel's spatial location. To preserve the structural information, we apply dense contrastive learning from multi-view correspondences. 

Given two views of an object, we can find corresponding pixels that project to the same location in 3D space. We require a view-invariant feature embedding -- a pixel should have high similarity with its corresponding pixel on the other image but remain different from all the other pixels. This encourages the model to learn the structural information of the object, to not collapse the embedding space, and to encode the object part consistently across different views. This multi-view contrastive objective only applies to two images of the same asset. However, because the underlying DINOv2 embedding space enables generalization for visually similar regions, the trained feature embedding can retain information about the structural similarities between functional parts \emph{across} categories. 

Formally, let $q$ be a pixel in the first view $I$, $q'_+$ be a pixel in the second view $I'$ that corresponds to the same location in 3D as $q$, and any other pixel on $I'$ be denoted as $q'_{-}$. The multiview contrastive objective is
\begin{equation}
    \mathcal{L}_{\text{spatial}} =  -\log \frac{e^{\text{sim}(q, q'_{+}) / \tau}}{e^{\text{sim}(q, q'_{+}) / \tau } + e^{\text{sim}(q, q'_{-}) / \tau}}.
\end{equation}
Combining the terms, we obtain the final loss
\begin{equation}
    \mathcal{L} = \mathcal{L}_\text{func}  + \lambda_\text{spatial} \mathcal{L}_\text{spatial} + \lambda_\text{mask} \mathcal{L}_\text{mask}.
\end{equation}

\input{figures_tex/training_objective}

\subsection{Implementation Details}

We use DINOv2-B as the backbone and an image size of 224. The MLP projector has 3 layers with 1024 hidden dimensions each. We use the Adam~\cite{kingma2017adammethodstochasticoptimization} optimizer with default hyperparameters, a batch size of 50 image pairs, 128 positive and negative sampled points on each image, and a learning rate of $1\times 10^{-4}$. In addition, we use a weight of $\lambda_\text{spatial} = 10$ for the spatial loss and a weight of $\lambda_\text{mask} = 1$ for the mask loss. We use random-color background augmentation during training following~\cite{florence2018dense}. A sensitivity analysis of loss weights and a breakdown of computational costs are provided in the Supplement.

%% file: figures_tex/architecture_diagram.tex
\begin{figure}[t]
    \centering
    \includegraphics[width=\linewidth]{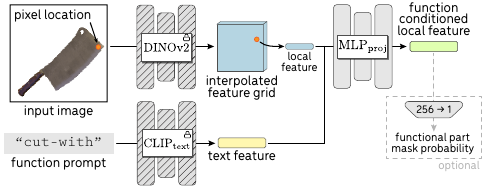}
    \vspace{-2em}
    \caption{\textbf{Local Functional Feature Extraction.} To obtain dense functionally conditioned features, we apply an MLP on top of a function text embedding and the spatial DINO features. The MLP is trained with both functional and spatial contrastive losses.}
    \vspace{-2em}
    \label{fig:architecture_diagram}
\end{figure}

%% file: figures_tex/training_objective.tex
\begin{figure}[t]
    \centering
    \includegraphics[width=\linewidth]{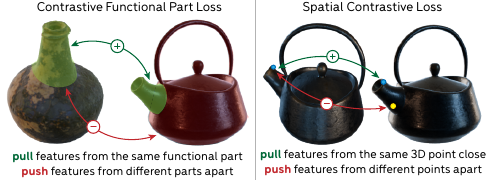}
    \vspace{-2em}
    \caption{\textbf{Training Objectives.} 
    To ensure functional part similarity in the learned feature space, we use a part-level contrastive objective to distill \emph{functional part semantics} from VLMs (left). The \emph{spatial} contrastive loss (right) serves a complementary role and prevents the model from collapsing predictions for different regions of a part, \eg, the top and bottom of a kettle spout.}
    \vspace{-1.5em}
    \label{fig:objective_figure}
\end{figure}

%% file: sec/5_experiments.tex
\section{Experiments}
\label{sec:experiments}

\input{table_tex/quantitative}

In this section, we benchmark our approach in \cref{sec:approach} and several baseline solutions on the dense functional correspondence task. Since our problem formulation in \cref{ssec:definition} requires a function as input and focuses on matches within functional parts, it differs significantly from existing benchmarks on semantic correspondence~\cite{min2019spair, sun2023misc210k}. As such, we leverage the evaluation datasets from \cref{sec:get-gt}.

\subsection{Metrics}
\label{sec:metrics}

We evaluate dense functional correspondence from two different aspects: \emph{correspondence label transfer}, which assesses the precision with which the model can transfer one functional part to another, and \emph{correspondence discovery}, which assesses the model's ability to identify relevant functional correspondences without any reference input labels.

\paragraph{Correspondence Label Transfer.}
To evaluate the precision of the correspondences that can be found using the learned features, we use normalized pixel distance (Normalized Dist) and percentage of correct keypoints (PCK).

Specifically, let the ground-truth correspondences between images $I_1, I_2$ given the function $\mathcal{F}$ be $\{p_1^1, p_1^2, \cdots, p_1^k\}, \{p_2^1, p_2^2, \cdots, p_2^k\}$. For each pixel $p_1^i$ on image $I_1$, we can find its most similar match $p_2^{j(i)}$ on $I_2$ using feature similarity. The normalized distance metric is simply the mean of $||p_2^{j(i)} - p_2^i||_2$ normalized by the image size, and PCK@k pixels is the mean of $\mathbbm{1}_{||p_2^{j(i)} - p_2^i||_2 < k}$.

\paragraph{Correspondence Discovery.}
In addition to label transfer, models should discover the relevant set of functional correspondences on its own, without assuming a priori that the relevant pixels on one image have been given. This capability is essential for potential downstream applications such as object alignment in robot object manipulation. 

First, since we assume that the input images are segmented, let $M_1, M_2$ be the object masks for images $I_1, I_2$. For every pixel $p_1^i \in M_1$, we find its most similar match $p_2^{j(i)}$ on $I_2$ and find the backward match of $p_2^{j(i)}$ on $I_1$, denoted as $q_1^i$. As such, $||p_1^i-q_1^i||_2$ captures the level of cycle-consistency of the match. We therefore construct a score $s = (1 - ||p_1^i-q_1^i||_2) \cdot \text{sim}(p_1^i, p_2^{j(i)})$ to rank each pair of $(p_1^i, p_2^{j(i)})$, using both similarity and cycle consistency. 

Then, we consider the top $t\%$ of all pairs as ``discovered'' and compare them with the ground-truth. A discovered pair $(x_1, x_2)$ is equivalent to a ground-truth pair $(y_1, y_2)$ if both end points are within $k$ pixels of the ground truth. Increasing $t$ results in higher recall but potentially lower precision: the number of discovered ground-truth correspondences monotonically increases while the percentage of correct correspondence tends to decrease. Sweeping $t$ produces a precision-recall curve, from which we can calculate the best F1 score (at $k$ pixels) and the average precision (AP) (at $k$ pixels). Formally, $\text{Best F1} = \max_{t} \frac{2\times \text{Precision}_t \times \text{Recall}_t}{\text{Precision}_t + \text{Recall}_t}$ and $\text{AP} = \sum_{t} (\text{Recall}_t - \text{Recall}_{t-1})\text{Precision}_t$.

\input{figures_tex/quali_discovery}

\subsection{Baselines}
\label{sec:baselines}

We describe several baseline methods below.

\paragraph{Self-Supervised Features.}
Powerful correspondences emerge in the feature space of large pre-trained vision foundation models, as reviewed in \cref{sec:lit-review}. 
We use features extracted from DINOv2~\cite{oquab2023dinov2}, Stable Diffusion~\cite{rombach2022high,zhang2023talefeaturesstablediffusion}, and fused features of the two~\cite{zhang2023talefeaturesstablediffusion} as baselines. We use feature-level similarity between pixel pairs to find correspondences.

\paragraph{Self-Supervised Features and VLM Grounding.} 
Since our task requires both semantic and structural reasoning based on the function prompt, these baselines chain a VLM that grounds functional parts with a pre-trained model that provides structural priors. Given an image pair, we use functional part bounding boxes generated by the VLM for each image, and then use self-supervised features to find correspondences within these part labels. 
This approach can benefit both label transfer and discovery because the functional part prediction adds a constraint on the space of possible matches, making it easier to find accurate matches. We consider two VLMs as the functional part grounding modules to be combined with off-the-shelf DINOv2 features:
\begin{itemize}
    \item CogVLM~\cite{wang2023cogvlm}, which outputs bounding boxes based on prompts of the functional part. 
    
    \item ManipVQA~\cite{huang2024manipvqa}, an affordance-grounding model that outputs bounding boxes conditioned on actions. We use the 7B model in our experiments. We also prompt ManipVQA in two ways, one with the functional part name and the other with the function itself because the model is finetuned for robotic tasks. We refer to these as ManipVQA-P and ManipVQA-F, respectively.
\end{itemize}

\subsection{Quantitative Comparisons}
\label{ssec:quantitative}

Results in \Cref{tab:quantitative} evaluate the performance of our method and baseline solutions on the synthetic and real evaluation datasets introduced in \cref{sec:get-gt}. Results show that our model trained on fully synthetic data can generalize to real images. 

Compared to baseline solutions that solely use self-supervised features, our full model -- trained with both functional and spatial contrastive loss -- consistently outperforms. These metrics demonstrate that the pseudo-label quality is sufficient for learning meaningful functional correspondences. Additionally, given that the evaluation dataset predominantly includes cross-category pairs, \Cref{tab:quantitative} illustrates that self-supervised features struggle with cross-category generalization. Further evidence is provided in the Supplement, where we present a detailed breakdown of metrics for both within- and across-category pairs.

Compared to baselines using VLM grounding, even with CogVLM bounding boxes as additional functional part information, off-the-shelf DINOv2 features underperform relative to our full model. The margin is generally smaller, which highlights the importance of understanding the context of the function. On the other hand, ManipVQA outputs less accurate bounding boxes, which is reflected in the metrics. In particular, prompting with the part instead of the function is significantly better, which shows the difficulty of zero-shot affordance grounding given a function name.
Note also that running CogVLM inference is roughly 50 times slower than our model and running ManipVQA inference is roughly 1000 times slower than our model.

\paragraph{Ablations.} We ablate the role of the functional and spatial contrastive loss in \Cref{tab:quantitative}. The model trained solely with functional loss performs poorly in both label transfer and correspondence discovery. The model trained solely with spatial loss is better but still falls short compared to the full model due to its lack of functional information.
Finally, models with and without mask loss share similar performances. The model with mask loss does outperform the model without it in all metrics for correspondence discovery on the real evaluation dataset, which represents the least constrained and most realistic case. This optional mask prediction module can learn functional part masks with minimal additional cost.

\subsection{Qualitative Results} 
\label{ssec:qualitative}

We present results for correspondence label transfer in \Cref{fig:intro_figure} and correspondence discovery in \Cref{fig:qualitative keypoint discovery}. Our model predictions not only capture object parts specific to the input function, but also preserve the structural relation among parts. \Cref{fig:qualitative keypoint discovery} shows top 10 matches according to the score from \Cref{sec:metrics} separated by 5 pixels each. DINOv2 features are not function-aware and result in inaccurate matching, especially in cross-category image pairs. 
In comparison, our model produces dense matches between functional parts from different object categories with high spatial precision, \eg, matching the rim of a saucepan with the rim of a jug. Overall, our model demonstrates a deep understanding of functional and structural information of objects, which produces high-quality dense functional correspondences.

%% file: table_tex/quantitative.tex
\begin{table*}[t]
  \centering
    \centering
    \footnotesize
    \setlength{\tabcolsep}{2pt}
    % \scalebox{0.86}{
    % \renewcommand{\arraystretch}{1.2}
        \begin{tabular}{lccccccc}
        \toprule
        \multirow{2}{*}{\textbf{Model}}&\multicolumn{3}{c}{\textbf{Correspondence Label Transfer}}&\multicolumn{4}{c}{\textbf{Correspondence Discovery}} \\
        \cmidrule(lr){2-4}\cmidrule(lr){5-8}
        & \makecell{Normalized Dist (↓)} & \makecell{PCK@23p ($\uparrow$)} & \makecell{PCK@10p ($\uparrow$)} & \makecell{Best F1@23p ($\uparrow$)} & \makecell{Best F1@10p ($\uparrow$)} & \makecell{AP@23p ($\uparrow$)} & \makecell{AP@10p ($\uparrow$)} \\
        \midrule
        % \textit{Synthetic Evaluation Dataset} \\[0.5em]
        \textit{Synthetic Evaluation Dataset} \\\midrule
        Chance & 0.310 & 0.165 & 0.046 & 0.416 & 0.176 & 0.256 & 0.093 \\
        \midrule
        DINO~\citep{oquab2023dinov2} & 0.212 & 0.381 & 0.148 & 0.578 & 0.281 & 0.381 & 0.130 \\
        SD~\citep{zhang2023talefeaturesstablediffusion} & 0.268 & 0.298 & 0.126 & 0.479 & 0.231 & 0.267 & 0.097 \\
        SD-DINO~\citep{zhang2023talefeaturesstablediffusion} & 0.227 & 0.376 & 0.161 & 0.563 & 0.301 & 0.341 & 0.144 \\
        \midrule
        CogVLM~\citep{wang2023cogvlm} + DINO & 0.180 & 0.416 & 0.158 & 0.678 & 0.333 & 0.556 & 0.188 \\
        ManipVQA-P~\citep{huang2024manipvqa} + DINO & 0.223 & 0.346 & 0.130 & 0.575 & 0.269 & 0.418 & 0.134 \\
        ManipVQA-F~\citep{huang2024manipvqa} + DINO & 0.272 & 0.259 & 0.093 & 0.528 & 0.244 & 0.320 & 0.097 \\
        \midrule
        Ours (functional only) & 0.228 & 0.287 & 0.094 & 0.575 & 0.233 & 0.441 & 0.112 \\
        Ours (spatial only) & 0.204 & 0.470 & \underline{0.227} & 0.610 & 0.369 & 0.412 & 0.211 \\
        Ours (full without mask loss) & \textbf{0.170} & \textbf{0.486} & \textbf{0.227} & \underline{0.768} & \underline{0.470} & \textbf{0.685} & \textbf{0.338} \\
        Ours (full with mask loss) & \underline{0.172} & \underline{0.480} & 0.223 & \textbf{0.774} & \textbf{0.471} & \underline{0.684} & \underline{0.330} \\\midrule
        % \underline{\textit{Real Evaluation Dataset}} \\[0.5em]
        \textit{Real Evaluation Dataset} \\\midrule
        Chance & 0.313 & 0.170 & 0.045 & 0.417 & 0.167 & 0.248 & 0.087 \\
        \midrule
        DINO~\citep{oquab2023dinov2} & 0.206 & 0.408 & 0.159 &  0.589 & 0.294 & 0.382 & 0.138 \\
        SD~\citep{zhang2023talefeaturesstablediffusion} & 0.259 & 0.309 & 0.127 & 0.503 & 0.238 & 0.285 & 0.101 \\
        SD-DINO~\citep{zhang2023talefeaturesstablediffusion} & 0.220 & 0.385 & 0.163 & 0.577 & 0.301 & 0.343 & 0.142 \\
        \midrule
        CogVLM~\citep{wang2023cogvlm} + DINO & 0.172 & 0.440 & 0.169 & 0.695 & 0.350 & 0.561 & 0.198 \\
        ManipVQA-P~\citep{huang2024manipvqa} + DINO & 0.204 & 0.398 & 0.153 & 0.600 & 0.295 & 0.420 & 0.148 \\
        ManipVQA-F~\citep{huang2024manipvqa} + DINO & 0.256 & 0.309 & 0.114 & 0.575 & 0.281 & 0.368 & 0.126 \\
        \midrule
        Ours (functional only) & 0.200 & 0.336 & 0.115 & 0.652 & 0.283 & 0.532 & 0.148 \\
        Ours (spatial only) & 0.203 & 0.472 & 0.228 & 0.708 & 0.353 & 0.382 & 0.182 \\
        Ours (full without mask loss) & \textbf{0.152} & \textbf{0.516} & \textbf{0.249} & \underline{0.775} & \underline{0.476} & \underline{0.691} & \underline{0.344} \\
        Ours (full with mask loss) & \underline{0.153} & \underline{0.501} & \underline{0.235} & \textbf{0.808} & \textbf{0.502} & \textbf{0.730} & \textbf{0.360} \\

        \bottomrule
    \end{tabular}
    \vspace{-5pt}
    \caption{\textbf{Quantitative Evaluation} on the synthetic and real evaluation datasets. The simplest baselines, self-supervised features from Stable Diffusion and DINOv2, perform relatively poorly. Adding semantic knowledge from predicted functional part labels from VLMs can offer slight improvement. Our approach, combining the strengths of both self-supervised features and VLMs, achieves the best performance.}
    \vspace{-1.5em}
  \label{tab:quantitative}
\end{table*}

%% file: figures_tex/quali_discovery.tex
\begin{figure*}[t]
    \centering
    \includegraphics[width=\linewidth]{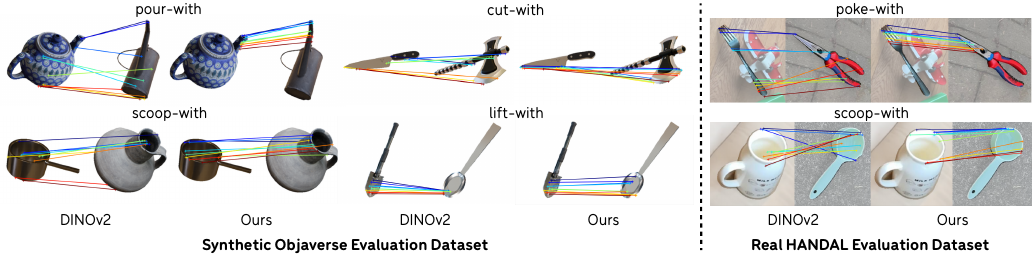}
    \vspace{-20pt}
    \caption{\textbf{Correspondence Discovery Comparisons.} We observe that our approach more reliably retrieves the functionally relevant correspondences than off-the-shelf DINOv2. The top 10 highest-ranked matches are shown.}
    \label{fig:qualitative keypoint discovery}
    \vspace{-1.5em}
\end{figure*}

%% file: sec/7_conclusion.tex
\section{Conclusion}

We have introduced the problem of dense functional correspondence, where input images contain objects with similar functionality but possibly come from distinct object categories. 
We have proposed a principled approach to obtain dense 2D functional correspondences from 3D object alignments and curated datasets for comprehensive evaluations. 
To tackle the task, we have presented a weakly-supervised framework that distills semantic information from vision-language models, while learning structural information through tuning self-supervised features with a multi-view contrastive loss. 
Our model outperforms a set of baselines in both synthetic and real-world benchmarks.

%% file: sec/8_supp_arxiv.tex
\section{Training Data Generation}

\subsection{Function and Object Taxonomy}
\label{sec:function_and_object_taxonomy}
\paragraph{List of functions.} To obtain our taxonomy of functions, we first take the function lists defined by~\cite{murali2021same, lai2021functional} and ask GPT-4~\cite{gpt4} to expand them. Our prompt is a simple ``Given this list of functions, generate more options for object functions.'' We manually process this list by simplifying synonymous functions into the most generic function to reduce redundancy, \eg, ``slice-with'' and  ``chop-with'' get absorbed into ``cut-with,'' or ``skewer-with'' and ``bore-with'' get absorbed into ``pierce-with.'' The final list of 24 functions is shown in \Cref{tab:functions_list}.

\paragraph{Finding object categories given a function.} To generate a list of object categories suitable for the chosen functions, we use the prompting strategy shown in~\Cref{fig:object_category_prompting}. We combine the common and uncommon lists and remove object names that are synonyms or that would require significant improvisation to achieve a certain function well.

\subsection{Functional Part Description Generation}
\label{subsec:functional_part_description}
Using LLMs, we have created a list of functions and a list of object categories that can carry out those functions. Given an object category and a function, we now require a means to generate part names and descriptions to prompt the grounded VLM. To obtain a list of functional part names, we use the prompting strategy shown in ~\Cref{fig:part_desc_gen}. This produces a list of parts for each \texttt{(object category, function)} pair, which we manually filter based on the most precise part. For instance, if GPT generates ``blade" and ``point" for \texttt{(knife, pierce-with)}, we will choose ``point.'' Querying different functions for the same object may result in the same functional part description being output multiple times with small variability. To combine these descriptions, we simply prompt GPT-4 to summarize them into one.

\subsection{Objaverse Dataset Filtering}
The Objaverse~\cite{deitke2023objaverse} dataset does not come with high-quality labels, making it challenging to use as a training dataset for tasks that require semantic object understanding. There is the Objaverse-LVIS split, but it is a small subset of the complete Objaverse, and the labels are noisy. To address this, Caption3D~\cite{luo2024caption3d} proposes a technique for generating captions for $\approx 600K$ of the assets in Objaverse based on a combination of VLMs and LLMs. However, these captions are still insufficient for our purpose because they do not contain explicit category labels. 

For each caption from Caption3D, we propose to filter it by comparing it with our list of object categories from \Cref{sec:function_and_object_taxonomy}. However, doing this naively using a large language model like Llama~\cite{dubey2024llama} would require about 100M model inferences, making this intractable. To resolve this and make the procedure more efficient, we propose summarizing each caption into a single noun using Llama3 with the prompting strategy described in~\Cref{fig:llama_summarization}. After converting the list of captions into a list of nouns, we use Llama3 word embedding distance to determine whether the noun belongs to the list of categories we generated in \Cref{sec:function_and_object_taxonomy}. Last, we ask Llama to verify the matches from word embedding as a final pass.

\input{figures_tex/pseudolabel_examples}

\subsection{CogVLM Prompting and Aggregation}
\label{subsec:cogvlm_labeling}

We use the descriptions generated in \Cref{subsec:functional_part_description} to prompt the \texttt{cogvlm-grounding-generalist-v1.1} variant of CogVLM~\cite{wang2023cogvlm}, which has been tuned for referring expression comprehension. Specifically, given a prompt like ``What are the exact bounding boxes of \texttt{<expr>} in the provided picture?'', where \texttt{<expr>} can be a noun or a descriptive phrase, the model is tuned to produce a text sequence describing a bounding box. Because of the sampling inherent to language transformer model inference, the bounding boxes vary across trials. Our procedure to label functional parts using CogVLM outputs consists of the following steps:

\begin{enumerate}
    \item Render 19 views per object that shows it from various angles, including from above and below. 
    \item For each functional part description and each view, query CogVLM for four trials to obtain the bounding box pseudo-labels. For small parts like points or tips, we do a second iteration that zooms into the initial bounding boxes to improve precision.
    \item Aggregate all trials and views onto a point cloud of 100K randomly sampled points on the object's surface. Every time a given point in the point cloud gets labeled by a bounding box in a different view, we increment its score. The final numbers are normalized to be in 0-1. For prompts that specifically ask for the labeling of edges, we multiply the point cloud with the per-point edge probabilities from SED-net~\cite{li2023surface}, a method for decomposing point clouds into primitives.
    \item Given this point cloud, for any rendered image of the object, we can project the point cloud into 2D and produce a binary mask with Otsu's method~\cite{otsu1975threshold} and a series of binary dilation/erosion steps to close holes in the mask.
\end{enumerate}

Example outputs of this procedure are shown in \Cref{fig:pseudolabel_examples}.

\input{table_tex/quantitative_breakdown}
\section{Additional Training and Evaluation Details}
\subsection{Ground-Truth Generation}

In this section, we provide additional details for deriving ground-truth 2D dense correspondences from 3D alignment. Given two object meshes that can perform the same function, we obtain their 3D functional alignment and the 3D bounding boxes for the functional parts using the procedure in \Cref{sec:get-gt}. Given rendered images $I_1, I_2$ of the two assets, we first find 2D pixels $P_1, P_2$ that would back-project to 3D points within the labeled functional part bounding boxes. The set of pixels $P_1$ and $P_2$ represent the functional part segmentation on the two images. Then, we perform minimum cost matching where the cost between two pixels $p_1 \in P_1$ and $p_2 \in P_2$ is measured by the distance between their back-projected 3D locations. In particular, we use the Hungarian algorithm. However, since the Hungarian algorithm requires one-to-one matches, we subsample the set between $P_1$ and $P_2$ that has more pixels using furthest point sampling. The output of the Hungarian algorithm constitutes the ground-truth 2D dense functional correspondences. 

Practically, we randomly sample rendered images from the top 5 out of 30 views where the functional part is most visible. We do so for six trials and repeat the procedure above to obtain 2D ground-truth annotations for the six view pairs for each pair of assets. Among these trials, ambiguity in the correspondence definition may arise due to 3D symmetries. We disambiguate this based on the objects' orientation when projected in the 2D images. For instance, for two rims in 2D, the top (in 2D, relative to the sides of the image) of one rim should align with the top of the other rim. We believe this is appropriate as it is the first investigation of this problem setting. In future work, we aim to refine the task and model to capture such ambiguity. As such, we manually filter the derived 2D annotations based on the ground-truth dense visualizations to disambiguate and ensure high quality.

\subsection{Additional Technical Details}

\paragraph{Model training.} For training our full models, we found that sampling points solely on the functional part for the spatial contrastive loss helped performance. However, when training the model with spatial loss only, we found that sampling points on the whole object helped more. 

\paragraph{Feature representation complexity.} We experimented with LoRA~\cite{hu2021loralowrankadaptationlarge} finetuning of DINOv2 and FiLM~\cite{perez2017filmvisualreasoninggeneral} layers for text conditioning. Despite the increased training cost of LoRA, we did not observe consistent improvements (e.g. normalized distance increased from 0.172 to 0.181).

\paragraph{Evaluation details.} Metrics are computed at fixed pixels because the input images are center-cropped (all objects have similar sizes), making it equivalent to normalizing with respect to a percentage of bounding box sizes as in prior work. For the SD-DINO baseline, we follow their standard resolutions and scale the input images accordingly. 

\paragraph{Evaluation with predicted functional part masks.} Below, we explain the evaluation protocol for models that involve a functional part mask prediction (\eg, CogVLM~\cite{wang2023cogvlm} + DINO, ManipVQA~\cite{huang2024manipvqa} + DINO, and our full model with mask loss). In label transfer, for each pixel $p_1^i$ on image $I_1$, we restrict its most similar match $p_2^{j(i)}$ on $I_2$ to be within the predicted functional part mask of $I_2$. In correspondence discovery, predicted part masks are produced for both images. We restrict matches to only happen between the two predicted masks and between their complements. Matches that fall within the two predicted part masks are prioritized in the ranking explained in \Cref{sec:metrics}.

\paragraph{Dense correspondence visualization.} The dense label transfer visualizations use the ground-truth mask for the source image but the predicted mask for the target image. For each pixel on the target image's functional part mask, we find its most similar match in the source image's functional part mask to produce the label transfer color map. 

\subsection{Computational Costs}

Rendering multi-view images on selected Objaverse~\cite{deitke2023objaverse} assets takes one day with four 2080 Ti GPUs. Functional part pseudo-labeling takes one week on eight A6000 GPUs, as CogVLM~\cite{wang2023cogvlm} inference is slow and memory-intensive. We emphasize that rendering and pseudo-labeling are only done once and scale significantly better than human annotation. Our model can be trained on a single NVIDIA GeForce RTX 3090 in $\approx2$ days for 100 epochs. These computational demands are fairly standard and are justified by the capability to trade off compute for expensive and time-consuming human annotation.

\section{Additional Quantitative Results}

\subsection{Within- and Cross-Category Comparison}

Since the evaluation dataset contains both within-category pairs and across-category pairs, we further separate the metrics in \Cref{tab:quantitative} into within-category results and across-category results in \Cref{tab:quantitative_breakdown}. In general, all the models and baselines perform better on within-category cases than on across-category cases. This illustrates the inherent difficulty of cross-category generalization. In addition, the performance margin between off-the-shelf self-supervised features and our model is often larger on the across-category pairs. On average, DINOv2 performs 46.3\% worse on cross-category pairs, while ours is 33.3\% worse. This serves as evidence that off-the-shelf self-supervised features struggle more with cross-category generalization. Last, without any functional part information, our spatial-only model performs competitively on within-category pairs on label transfer metrics but is worse on across-category pairs.

\input{table_tex/quantitative_scaling}
\input{table_tex/weight_ablation}
\input{table_tex/quantitative_iou}

\input{figures_tex/supp_quali_dense}

\subsection{Scaling Experiments}

In this section, we show scaling experiments where we replace the backbone in our full model with mask loss with DINOv2~\cite{oquab2023dinov2} of different ViT sizes. The results are shown in \Cref{tab:quantitative_scaling}. As the ViT size increases, we generally observe an improvement in the evaluation metrics. In addition, when we reduce the stride size from 14 pixels to 7, the model performance also improves, especially in correspondence discovery. This demonstrates that both higher spatial resolution and higher backbone capacity can improve the performance of our approach.

Note that due to computational resource constraints, DINOv2 with ViT-G was only trained for 30 epochs, and ViT-B with half stride was trained for 80 epochs, while other models were trained for 100 epochs. Compared to ViT-B, using ViT-S is $\approx1.6$ times faster, using half stride is $\approx2.6$ times slower, using ViT-L is $\approx2.1$ times slower, and using ViT-G is $\approx5.8$ times slower.

\subsection{Sensitivity Analysis of Loss Weights}

We further ablate the spatial and mask loss weights in \Cref{tab:weight_ablation}. Varying $\lambda_{\text{spatial}}$ has an effect, but the model does not appear to be highly sensitive, making it easy to converge on $\lambda_{\text{spatial}}=10$ to achieve the best result. On the other hand, we observe low variance when increasing $\lambda_{\text{mask}}$. The benefits of the mask loss are illustrated in \Cref{fig:mask_loss_comparison}.

\subsection{Functional Part Prediction Accuracy}

Some of the methods we evaluate generate functional part segmentation predictions. Accordingly, we compare their segmentation accuracies in \Cref{tab:quantitative_iou}. Specifically, ManipVQA-P and ManipVQA-F~\citep{huang2024manipvqa} refer to segmentations produced by ManipVQA using part label prompts and function name prompts, respectively. For CogVLM~\citep{wang2023cogvlm} on 2D images, predictions are generated from single-image inputs into CogVLM, aggregated across four trials via K-Means clustering. These three methods produce bounding boxes, which are further multiplied with the object masks. CogVLM~\citep{wang2023cogvlm} with 3D aggregation follows the pipeline illustrated in \Cref{fig:labeling_pipeline}. Since our full model with mask loss incorporates a functional part mask prediction module, we also evaluate its segmentation performance as part of this comparison. 

To evaluate these methods, we use ground-truth part masks generated by our evaluation pipeline on both the synthetic Objaverse~\cite{deitke2023objaverse} data and the real HANDAL~\cite{guo2023handal} data. Specifically, for each \texttt{(object, function)} pair, we label a 3D bounding box, and any pixel that projects to a 3D point within this bounding box is classified as part of the functional region. As shown in \Cref{tab:quantitative_iou}, both CogVLM methods and our learned model have good accuracy. Note that the pseudo-labeling pipeline can produce very fine-grained parts like small tips or edges that do not necessarily align with the human annotations. As such, the main advantage of the 3D aggregation pipeline is illustrated in \Cref{fig:pseudolabel_examples}. In addition, on the real HANDAL data, our model's predictions perform better than the CogVLM model, which has state-of-the-art referring expression detection capabilities.

\subsection{Ranking Scheme.} We designed our feature similarity and cycle consistency-based ranking scheme in \Cref{sec:metrics} to enable strong performance across all methods. To show its impact, we include results from a simpler version using only feature similarity in \Cref{tab:sim_only_ranking}. The ordering is consistent with the main text, but all methods perform worse. This confirms that all methods benefit from the improved ranking scheme and that our findings are not sensitive to this.

\input{table_tex/sim_only_ranking}

\section{Additional Qualitative Results}

Additional dense label transfer results on the synthetic Objaverse dataset, which further validate the effectiveness of our approach, are presented in \Cref{fig:addnl_dense}. These results highlight the strong performance of our model in transferring functional part labels across diverse object categories.

More qualitative discovery results on the synthetic Objaverse dataset are shown in \Cref{fig:addnl_quali_objaverse}, and more qualitative discovery results on the real HANDAL dataset are shown in \Cref{fig:addnl_quali_handal}. We compare our model with the DINO~\cite{oquab2023dinov2} and CogVLM~\cite{wang2023cogvlm} + DINO baselines. In line with the conclusion in \Cref{ssec:qualitative}, our model can focus on the functionally relevant part and produce more spatially precise correspondences. 

Lastly, we show qualitative evidence for the potential benefits of the optional mask loss in \Cref{fig:mask_loss_comparison}. In cases where the predicted functional part mask is accurate, the mask loss can prevent incorrect matches outside functionally relevant regions.

\input{figures_tex/mask_loss_comparison_figure}

\section{Discussion}

\paragraph{Differences with FunKPoint~\cite{lai2021functional}.} The concept of functional correspondence was previously introduced by~\cite{lai2021functional}. However, our formulation is different in three key aspects.

First, our problem requires dense functional correspondences to be established, whereas \cite{lai2021functional} defines five sparse keypoints. The manual definition of semantic keypoints at the function type level is not guaranteed to be well-defined across all object categories. Consequently, we observe inconsistencies and labeling ambiguities in the sparse keypoint annotations. In addition, establishing dense correspondences requires fine-grained and precise reasoning about the structure of object parts, which may make it more useful for downstream applications like transferring demonstrations in robotics.

Second, keypoint matches from \cite{lai2021functional} include both the object's functional part and where the human interacts with the object. In many cases, like a ``bottle" and a ``kettle," the functionally irrelevant parts cannot be well aligned. As a result, the key points outside of the functional parts are highly ambiguous. In addition, where an agent interacts with the object depends on the end-effector design, introducing complexity and confounding. On the other hand, our formulation introduces a more precise definition based on the concept of functionally equivalent 3D alignments (discussed in \Cref{sec:problem-definition}).

Last, the model proposed by \citet{lai2021functional} relies on human annotations of sparse keypoints, which inherently limits scalability. In contrast, our approach leverages self-supervised features and pseudo-labeling, requiring minimal human input, and offers a significantly more scalable solution.

Because of these fundamental differences, our method is not directly applicable to the dataset in \citet{lai2021functional}. While the feature maps from \citet{lai2021functional} could be used for dense correspondence, the method is not designed for this and it qualitatively appears to be relatively coarse. A visual comparison is provided in \Cref{fig:comparison_with_functional}.

\input{figures_tex/comparison_with_functional_paper}

\textbf{Limitations.} A limitation of our work is the existence of ambiguity in some cross-category cases. Ambiguity can arise when an object has multiple parts that can be used for the same function. For instance, both the tip and side rim of a spoon can be used for the function ``scrape-with." On the other hand, ambiguity can also arise due to radial symmetry: the rim of a cup and the rim of a bowl can be matched in infinitely many ways. As such, a compelling direction for future work can be developing a probabilistic model to handle the multi-modal nature of the problem and use additional conditioning to resolve such ambiguities.

\clearpage
\input{figures_tex/supp_quali_objaverse}
\input{figures_tex/supp_quali_handal}

\clearpage
\input{figures_tex/object_category_prompting}
\input{figures_tex/part_desc_gen}
\input{figures_tex/llama_summarization}

\clearpage
\input{table_tex/taxonomy}

\clearpage
\input{table_tex/object_counts}

%% file: figures_tex/pseudolabel_examples.tex
\begin{figure*}[t]
    \centering
    \includegraphics[width=\linewidth]{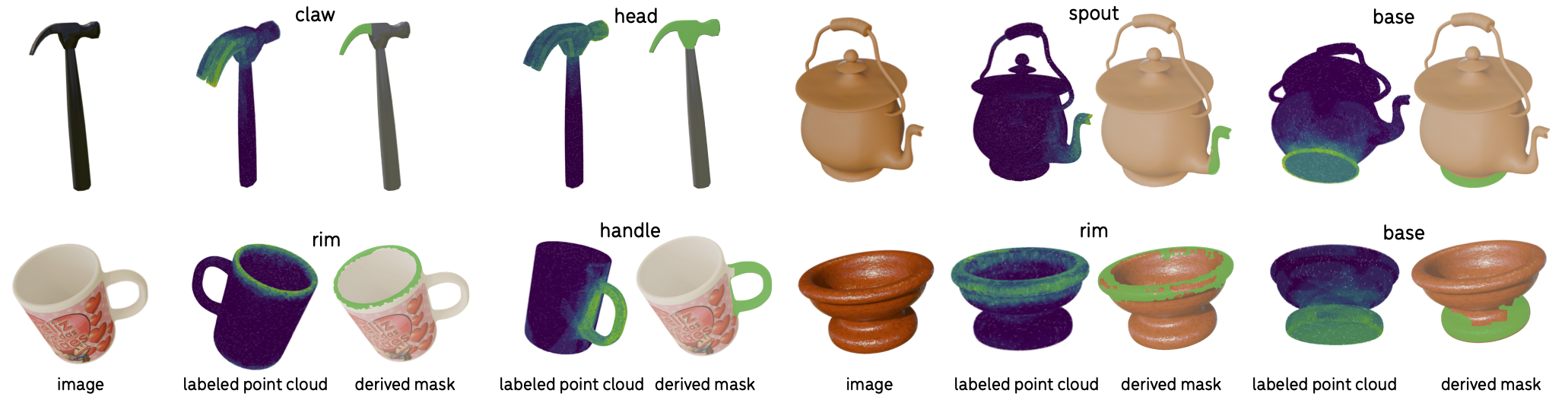}
    \caption{\textbf{Examples of pseudo-labeled functional parts in point clouds and images using CogVLM~\cite{wang2023cogvlm}.} Using the procedure outlined in \Cref{subsec:cogvlm_labeling}, we pseudo-label images with masks for the object functional parts. Notably, this pipeline has the ability to generate part labels for non-convex object parts, such as a mug's rim, and for parts that lack clear edge boundaries, such as a teapot's spout. Point clouds are shown in views that best capture the aggregated functional part labels.}
    \label{fig:pseudolabel_examples}
\end{figure*}

%% file: table_tex/quantitative_breakdown.tex
\begin{table*}[t]
  \centering
    \centering
    \footnotesize
    \setlength{\tabcolsep}{2pt}
    \scalebox{1.0}{
        \begin{tabular}{lccccccc}
        \toprule
        \multirow{2}{*}{\textbf{Model}}&\multicolumn{3}{c}{\textbf{Correspondence Label Transfer}}&\multicolumn{4}{c}{\textbf{Correspondence Discovery}} \\
        \cmidrule(lr){2-4}\cmidrule(lr){5-8}
        & \makecell{Normalized Dist (↓)} & \makecell{PCK@23p ($\uparrow$)} & \makecell{PCK@10p ($\uparrow$)} & \makecell{Best F1@23p ($\uparrow$)} & \makecell{Best F1@10p ($\uparrow$)} & \makecell{AP@23p ($\uparrow$)} & \makecell{AP@10p ($\uparrow$)} \\
        & \makecell{within / across} & \makecell{within / across} & \makecell{within / across} & \makecell{within / across} & \makecell{within / across} & \makecell{within / across} & \makecell{within / across} \\
        \midrule
        
        % \textit{Synthetic Evaluation Dataset} \\[0.5em]
        \multicolumn{2}{l}{\textit{Synthetic Evaluation Dataset}} \\ \midrule
        Chance & 0.317 / 0.309 & 0.162 / 0.166 & 0.047 / 0.046 & 0.382 / 0.421 & 0.163 / 0.178 & 0.234 / 0.260 & 0.085 / 0.095 \\
        \midrule
        DINO~\citep{oquab2023dinov2} & 0.132 / 0.225 & 0.589 / 0.347 & 0.283 / 0.126 & 0.708 / 0.557 & 0.425 / 0.257 & 0.555 / 0.352 & 0.265 / 0.108\\
        SD~\citep{zhang2023talefeaturesstablediffusion} & 0.221 / 0.275 & 0.423 / 0.278 & 0.210 / 0.112 & 0.528 / 0.471 & 0.295 / 0.220 & 0.322 / 0.258 & 0.153 / 0.087\\
        SD-DINO~\citep{zhang2023talefeaturesstablediffusion} & 0.154 / 0.240 & 0.553 / 0.347 & 0.284 / 0.141 & 0.642 / 0.550 & 0.406 / 0.284 & 0.443 / 0.324 & 0.239 / 0.129 \\
        \midrule
        CogVLM~\citep{wang2023cogvlm} + DINO & 0.126 / 0.188 & 0.596 / 0.387 & 0.281 / 0.138 & 0.840 / 0.651 & 0.519 / 0.303 & 0.749 / 0.525 & 0.362 / 0.160\\
        CogVLM~\citep{wang2023cogvlm} + SD-DINO & 0.135 / 0.188 & 0.578 / 0.404 & 0.292 / 0.161 & 0.825 / 0.683 & 0.554 / 0.368 & 0.717 / 0.551 & 0.400 / 0.216\\
        ManipVQA-P~\citep{huang2024manipvqa} + DINO & 0.181 / 0.230 & 0.493 / 0.323 & 0.232 / 0.113 & 0.737 / 0.548 & 0.437 / 0.242 & 0.608 / 0.387 & 0.284 / 0.110 \\
        ManipVQA-F~\citep{huang2024manipvqa} + DINO & 0.234 / 0.278 & 0.352 / 0.244 & 0.152 / 0.084 & 0.650 / 0.508 & 0.374 / 0.222 & 0.444 / 0.300 & 0.193 / 0.081 \\
        \midrule
        Ours (functional only) & 0.187 / 0.235 & 0.412 / 0.266 & 0.154 / 0.084 & 0.723 / 0.551 & 0.358 / 0.212 & 0.617 / 0.412 & 0.220 / 0.094 \\
        Ours (spatial only) & 0.128 / 0.217 & \underline{0.674} / 0.436 & \textbf{0.385} / \underline{0.201} & 0.686 / 0.597 & 0.469 / 0.353 & 0.493 / 0.398 & 0.295 / 0.198\\
        Ours (full without mask loss) & \textbf{0.112} / \textbf{0.180} & \textbf{0.680} / \textbf{0.454} & \underline{0.377} / \textbf{0.203} & \underline{0.878} / \underline{0.750} & \underline{0.643} / \underline{0.442} & \textbf{0.823} / \textbf{0.662} & \textbf{0.537} / \textbf{0.306} \\
        Ours (full with mask loss) & \underline{0.122} / \underline{0.180} & 0.655 / \underline{0.451} & 0.367 / 0.199 & \textbf{0.879} / \textbf{0.757} & \textbf{0.645} / \textbf{0.443} & \underline{0.820} / \underline{0.661} & \underline{0.528} / \underline{0.297} \\\midrule
        % \underline{\textit{Real Evaluation Dataset}} \\[0.5em]
        \multicolumn{2}{l}{\textit{Real Evaluation Dataset}} \\\midrule
        Chance & 0.311 / 0.313 & 0.170 / 0.170 & 0.044 / 0.046 & 0.431 / 0.413 & 0.171 / 0.165 & 0.262 / 0.243 & 0.090 / 0.086\\
        \midrule
        DINO~\citep{oquab2023dinov2} & 0.130 / 0.230 & 0.570 / 0.356 & 0.252 / 0.129 & 0.734 / 0.542 & 0.434 / 0.250 & 0.577 / 0.320 & 0.275 / 0.095\\
        SD~\citep{zhang2023talefeaturesstablediffusion} & 0.204 / 0.277 & 0.411 / 0.276 & 0.192 / 0.106 & 0.587 / 0.477 & 0.308 / 0.215 & 0.355 / 0.263 & 0.148 / 0.086 \\
        SD-DINO~\citep{zhang2023talefeaturesstablediffusion} & 0.151 / 0.243 & 0.514 / 0.344 & 0.244 / 0.137 & 0.679 / 0.544 & 0.400 / 0.270 & 0.468 / 0.303 & 0.224 / 0.116 \\
        \midrule
        CogVLM~\citep{wang2023cogvlm} + DINO & 0.142 / 0.182 & 0.544 / 0.407 & 0.239 / 0.147 & 0.782 / 0.667 & 0.465 / 0.314 & 0.686 / 0.521 & 0.312 / 0.161 \\
        CogVLM~\citep{wang2023cogvlm} + SD-DINO & 0.154 / 0.186 & 0.506 / 0.402 & 0.234 / 0.158 & 0.762 / 0.683 & 0.462 / 0.360 & 0.618 / 0.540 & 0.295 / 0.219 \\
        ManipVQA-P~\citep{huang2024manipvqa} + DINO & 0.148 / 0.222 & 0.534 / 0.354 & 0.234 / 0.127 & 0.719 / 0.563 & 0.415 / 0.256 & 0.577 / 0.370 & 0.260 / 0.112 \\
        ManipVQA-F~\citep{huang2024manipvqa} + DINO & 0.236 / 0.263 & 0.405 / 0.279 & 0.174 / 0.095 & 0.714 / 0.531 & 0.412 / 0.239 & 0.509 / 0.323 & 0.231 / 0.093 \\
        \midrule
        Ours (functional only) & 0.179 / 0.206 & 0.405 / 0.313 & 0.152 / 0.103 & 0.730 / 0.627 & 0.356 / 0.260 & 0.599 / 0.511 & 0.199 / 0.132 \\
        Ours (spatial only) & \underline{0.129} / 0.227 & \underline{0.631} / 0.421 & \underline{0.343} / 0.192 & 0.708 / 0.564 & 0.470 / 0.316 & 0.501 / 0.344 & 0.295 / 0.145 \\
        Ours (full without mask loss) & \textbf{0.122} / \underline{0.161} & \textbf{0.639} / \textbf{0.477} & \textbf{0.352} / \textbf{0.216} & \underline{0.835} / \underline{0.756} & \underline{0.589} / \underline{0.441} & \underline{0.741} / \underline{0.675} & \underline{0.469} / \underline{0.304} \\
        Ours (full with mask loss) & 0.132 / \textbf{0.160} & 0.603 / \underline{0.469} & 0.321 / \underline{0.208} & \textbf{0.857} / \textbf{0.792} & \textbf{0.611} / \textbf{0.467} & \textbf{0.773} / \textbf{0.716} & \textbf{0.485} / \textbf{0.321} \\
        \bottomrule
    \end{tabular}
    }
    \vspace{-5pt}
    \caption{\textbf{Quantitative evaluation by within- and across-category pairs.} We further break down \Cref{tab:quantitative} by within- and across-category performance for all the metrics. Additional result for CogVLM + SD-DINO is also included. Off-the-shelf self-supervised features tend to perform worse at cross-category generalization compared to our full model.
    }
    %\vspace{-1.5em}
  \label{tab:quantitative_breakdown}
\end{table*}

%% file: table_tex/quantitative_scaling.tex
\begin{table*}[t]
  \centering
    \centering
    \footnotesize
    \setlength{\tabcolsep}{2pt}
    \scalebox{1.0}{
        \begin{tabular}{lccccccc}
        \toprule
        \multirow{2}{*}{\textbf{Model}}&\multicolumn{3}{c}{\textbf{Correspondence Label Transfer}}&\multicolumn{4}{c}{\textbf{Correspondence Discovery}} \\
        \cmidrule(lr){2-4}\cmidrule(lr){5-8}
        & \makecell{Normalized Dist (↓)} & \makecell{PCK@23p ($\uparrow$)} & \makecell{PCK@10p ($\uparrow$)} & \makecell{Best F1@23p ($\uparrow$)} & \makecell{Best F1@10p ($\uparrow$)} & \makecell{AP@23p ($\uparrow$)} & \makecell{AP@10p ($\uparrow$)} \\
        \midrule
        % \textit{Synthetic Evaluation Dataset} \\[0.5em]
        \multicolumn{2}{l}{\textit{Synthetic Evaluation Dataset}} \\ \midrule
        DINOv2 ViT-S & 0.171 & 0.476 & 0.218 & 0.768 & 0.466 & 0.676 & 0.325 \\
        DINOv2 ViT-B & 0.172 & 0.480 & 0.223 & 0.774 & 0.471 & 0.684 & 0.330 \\        
        DINOv2 ViT-B w/ half stride & 0.166 & \underline{0.494} & 0.229 & \textbf{0.799} & \textbf{0.508} & \textbf{0.721} & \textbf{0.373} \\
        DINOv2 ViT-L & \underline{0.164} & 0.493 & \underline{0.233} & 0.789 & 0.490 & 0.705 & 0.351 \\
        DINOv2 ViT-G & \textbf{0.161} & \textbf{0.505} & \textbf{0.239} & \underline{0.792} & \underline{0.498} & \underline{0.711} & \underline{0.361}\\
        \midrule
        % \underline{\textit{Real Evaluation Dataset}} \\[0.5em]
        \multicolumn{2}{l}{\textit{Real Evaluation Dataset}} \\\midrule
        DINOv2 ViT-S & 0.162 & 0.494 & 0.229 & 0.788 & 0.481 & 0.697 & 0.335\\
        DINOv2 ViT-B & 0.153 & 0.501 & 0.235 & 0.808 & 0.502 & 0.730 & 0.360 \\
        DINOv2 ViT-B w/ half stride & \underline{0.150} & \underline{0.519} & \underline{0.247} & \textbf{0.821} & \textbf{0.525} & \textbf{0.751} & \textbf{0.403} \\
        DINOv2 ViT-L & 0.152 & 0.515 & 0.244 & \underline{0.809} & \underline{0.514} & \underline{0.730} & \underline{0.377} \\
        DINOv2 ViT-G & \textbf{0.146} & \textbf{0.523} & \textbf{0.252} & 0.808 & 0.507 & 0.729 & 0.370\\
        \bottomrule
    \end{tabular}
    }
    \vspace{-5pt}
    \caption{\textbf{Quantitative evaluation of our model trained with different backbones.} In general, performance increases when the vision transformer backbone becomes larger or when the stride size is reduced.
    }
    %\vspace{-1.5em}
  \label{tab:quantitative_scaling}
\end{table*}

%% file: table_tex/weight_ablation.tex
\begin{table*}[t]
  \centering
    \centering
    \footnotesize
    \setlength{\tabcolsep}{2pt}
    \scalebox{1.0}{
    \begin{tabular}{lccccccc}
    \toprule
    \multirow{2}{*}{\textbf{Loss Weights}}&\multicolumn{3}{c}{\textbf{Correspondence Label Transfer}}&\multicolumn{4}{c}{\textbf{Correspondence Discovery}} \\
    \cmidrule(lr){2-4}\cmidrule(lr){5-8}
    & \makecell{Normalized Dist (↓)} & \makecell{PCK@23p ($\uparrow$)} & \makecell{PCK@10p ($\uparrow$)} & \makecell{Best F1@23p ($\uparrow$)} & \makecell{Best F1@10p ($\uparrow$)} & \makecell{AP@23p ($\uparrow$)} & \makecell{AP@10p ($\uparrow$)} \\
    \midrule
    % \textit{Synthetic Evaluation Dataset} \\[0.5em]
    \multicolumn{2}{l}{\textit{Synthetic Evaluation Dataset}} \\ \midrule
    $\lambda_{\text{spatial}} = 1$, $\lambda_{\text{mask}} = 1$ & 0.193 & 0.402 & 0.161 & 0.707 & 0.367 & 0.601 & 0.224 \\
    $\lambda_{\text{spatial}} = 5$, $\lambda_{\text{mask}} = 1$ & 0.177 & 0.458 & 0.207 & 0.761 & 0.445 & 0.664 & 0.304\\
    $\lambda_{\text{spatial}} = 10$, $\lambda_{\text{mask}} = 1$ & \underline{0.172} & \textbf{0.480} & \textbf{0.223} & 0.774 & \textbf{0.471} & 0.684 & \underline{0.330} \\
    $\lambda_{\text{spatial}} = 10$, $\lambda_{\text{mask}} = 5$ & 0.173 & 0.477 & \underline{0.222} & \underline{0.775} & \underline{0.471} & \underline{0.685} & \textbf{0.330} \\
    $\lambda_{\text{spatial}} = 10$, $\lambda_{\text{mask}} = 10$ & \textbf{0.170} & \underline{0.478} & 0.221 & \textbf{0.778} & 0.470 & \textbf{0.687} & 0.329 \\
    \midrule
    % \underline{\textit{Real Evaluation Dataset}} \\[0.5em]
    \multicolumn{2}{l}{\textit{Real Evaluation Dataset}} \\\midrule
    $\lambda_{\text{spatial}} = 1$, $\lambda_{\text{mask}} = 1$ & 0.169 & 0.443 & 0.175 & 0.759 & 0.405 & 0.671 & 0.260 \\
    $\lambda_{\text{spatial}} = 5$, $\lambda_{\text{mask}} = 1$ & 0.158 & 0.492 & 0.223 & 0.793 & 0.481 & 0.713 & 0.345 \\
    $\lambda_{\text{spatial}} = 10$, $\lambda_{\text{mask}} = 1$ & \textbf{0.153} & \underline{0.501} & \textbf{0.235} & \textbf{0.808} & \textbf{0.502} &\textbf{ 0.730} & \textbf{0.360} \\
    $\lambda_{\text{spatial}} = 10$, $\lambda_{\text{mask}} = 5$ & 0.155 & 0.499 & \underline{0.232} & \underline{0.804} & \underline{0.497} & \underline{0.729} & \underline{0.359} \\
    $\lambda_{\text{spatial}} = 10$, $\lambda_{\text{mask}} = 10$ & \underline{0.154} & \textbf{0.501} & 0.231 & 0.800 & 0.495 & 0.719 & 0.353 \\
    \bottomrule
    \end{tabular}
    }
    \vspace{-5pt}
    \caption{\textbf{Quantitative evaluation of varying loss weights.} Model performance improves with increasing spatial loss weight (up to 10) and remains stable with different mask loss weights.}
    %\vspace{-3.0mm}
  \label{tab:weight_ablation}
\end{table*}

%% file: table_tex/quantitative_iou.tex
\begin{table}[t]
  \centering
    \centering
    \footnotesize
    \setlength{\tabcolsep}{2pt}
    %\scalebox{0.88}{
    % \renewcommand{\arraystretch}{1.2}
        \begin{tabular}{lcc}
        \toprule
        \multirow{1}{*}{\textbf{Method}} & IoU on Objaverse & IoU on HANDAL\\
        \midrule
        ManipVQA-P~\citep{huang2024manipvqa} & 0.453 & 0.276 \\
        ManipVQA-F~\citep{huang2024manipvqa} & 0.240 & 0.146 \\
        CogVLM~\citep{wang2023cogvlm} on 2D images & \textbf{0.656} & \underline{0.597}  \\
        CogVLM~\citep{wang2023cogvlm} w/ 3D aggregation & \underline{0.635} & N/A\\
        Our model prediction & 0.628 & \textbf{0.636}\\
        \bottomrule
    \end{tabular}
    %}
    \vspace{-5pt}
    \caption{\textbf{Quantitative evaluation of functional part segmentation accuracy.} This table compares the accuracy of functional part segmentation produced by different methods. Both CogVLM~\citep{wang2023cogvlm} and the predictions of the distilled model demonstrate strong performance in this task. Note that the pipeline described in Figure 3 in the main paper was applied only to the Objaverse dataset; therefore, results for CogVLM~\citep{wang2023cogvlm} with 3D aggregation are omitted for the HANDAL dataset.}
    %\vspace{-1.5em}
  \label{tab:quantitative_iou}
\end{table}

%% file: figures_tex/supp_quali_dense.tex
\begin{figure*}[t]
    \centering
    \includegraphics[width=\linewidth]{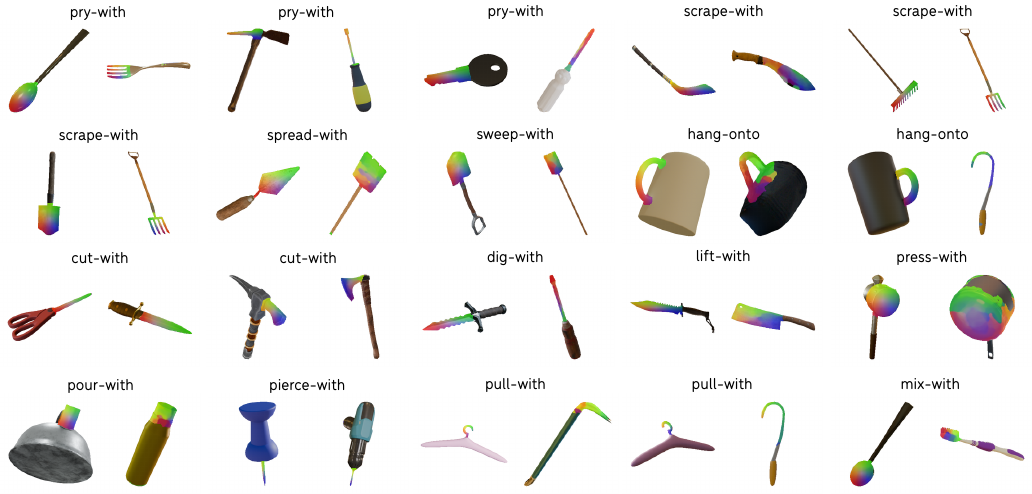}
    \caption{\textbf{Additional Label Transfer Dense Visualizations.} For each target image (right), our model predicts the functional part mask. To generate the transferred color map, each pixel in the predicted mask is matched to its best corresponding pixel within the ground-truth mask of the source image (left) in terms of feature similarity.}
    \label{fig:addnl_dense}
\end{figure*}

%% file: table_tex/sim_only_ranking.tex
\begin{table}[t]
  \centering
    \centering
    \footnotesize
    \setlength{\tabcolsep}{2pt}
    \scalebox{0.8}{
        \begin{tabular}{lcccc}
        \toprule
        \textbf{Model} & \makecell{Best F1@23p ($\uparrow$)} & \makecell{Best F1@10p ($\uparrow$)} & \makecell{AP@23p ($\uparrow$)} & \makecell{AP@10p ($\uparrow$)} \\
        \midrule
        DINO & 0.573 & 0.277 & 0.376 & 0.128 \\
        CogVLM + DINO & 0.672 & 0.329 & 0.551 & 0.184 \\
        Ours (full with mask loss) & 0.767 & 0.465 & 0.679 & 0.325 \\
        \bottomrule
    \end{tabular}
    }
    \vspace{-5pt}
    \caption{\textbf{Correspondence discovery evaluation using only feature similarity.} Compared with \Cref{tab:quantitative}, using only feature similarity in the ranking scheme achieves worse performance overall but preserves relative performance among methods.}
  \label{tab:sim_only_ranking}
\end{table}
% \vspace{-2.5mm}

%% file: figures_tex/mask_loss_comparison_figure.tex
\begin{figure}[t]
    \vspace{-3mm}
    \centering
    \includegraphics[width=\linewidth]{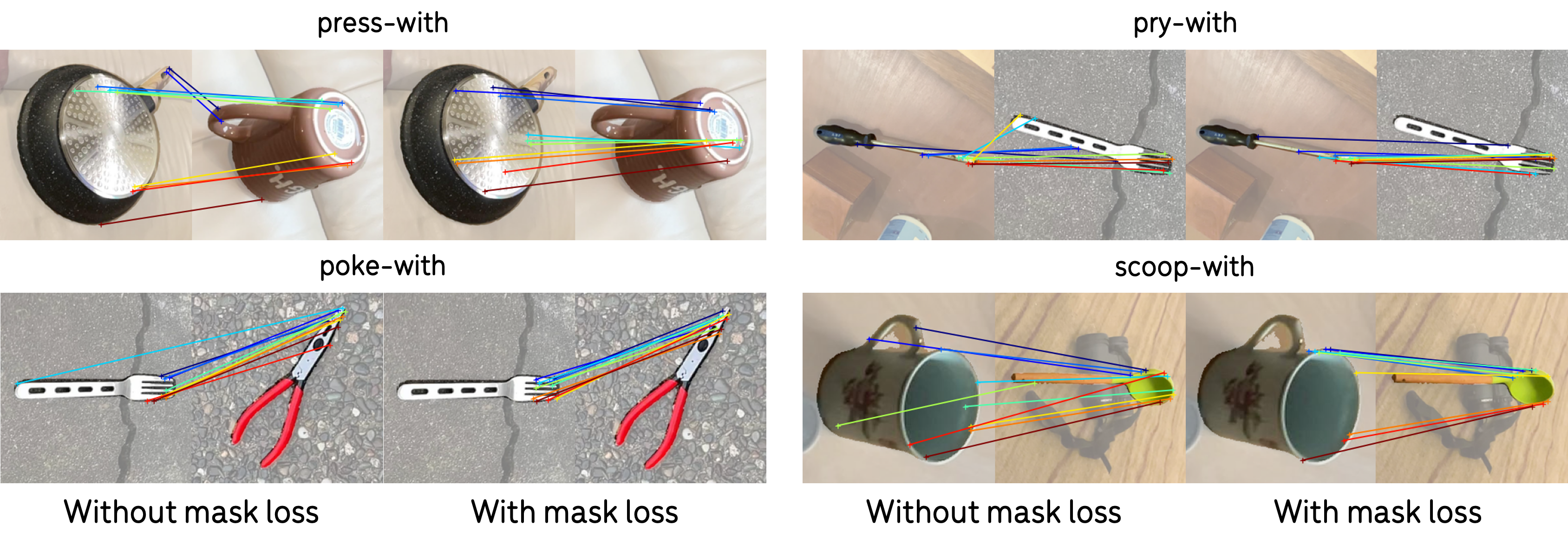}
    \vspace{-2em}
    \caption{\textbf{Qualitative examples for the impact of mask loss.} Functional part predictions can help avoid incorrect matches outside the functional parts in correspondence discovery.}
    \vspace{-1.5em}
    \label{fig:mask_loss_comparison}
\end{figure}

%% file: figures_tex/comparison_with_functional_paper.tex
\begin{figure}[t]
    \vspace{3mm}
    \centering
    \includegraphics[width=\linewidth]{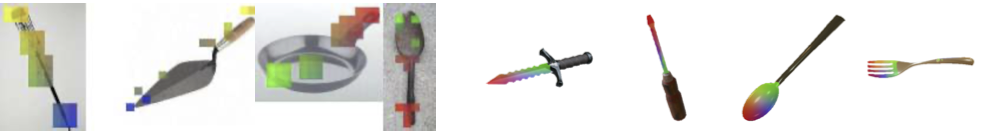}
    \vspace{-2em}
    \caption{\textbf{Comparison with \cite{lai2021functional}} A visual comparison of dense correspondence between \cite{lai2021functional} (left) and our method (right).}
    \vspace{-1.5em}
    \label{fig:comparison_with_functional}
\end{figure}

%% file: figures_tex/supp_quali_objaverse.tex
\begin{figure*}[t]
    \centering
    \includegraphics[width=\linewidth]{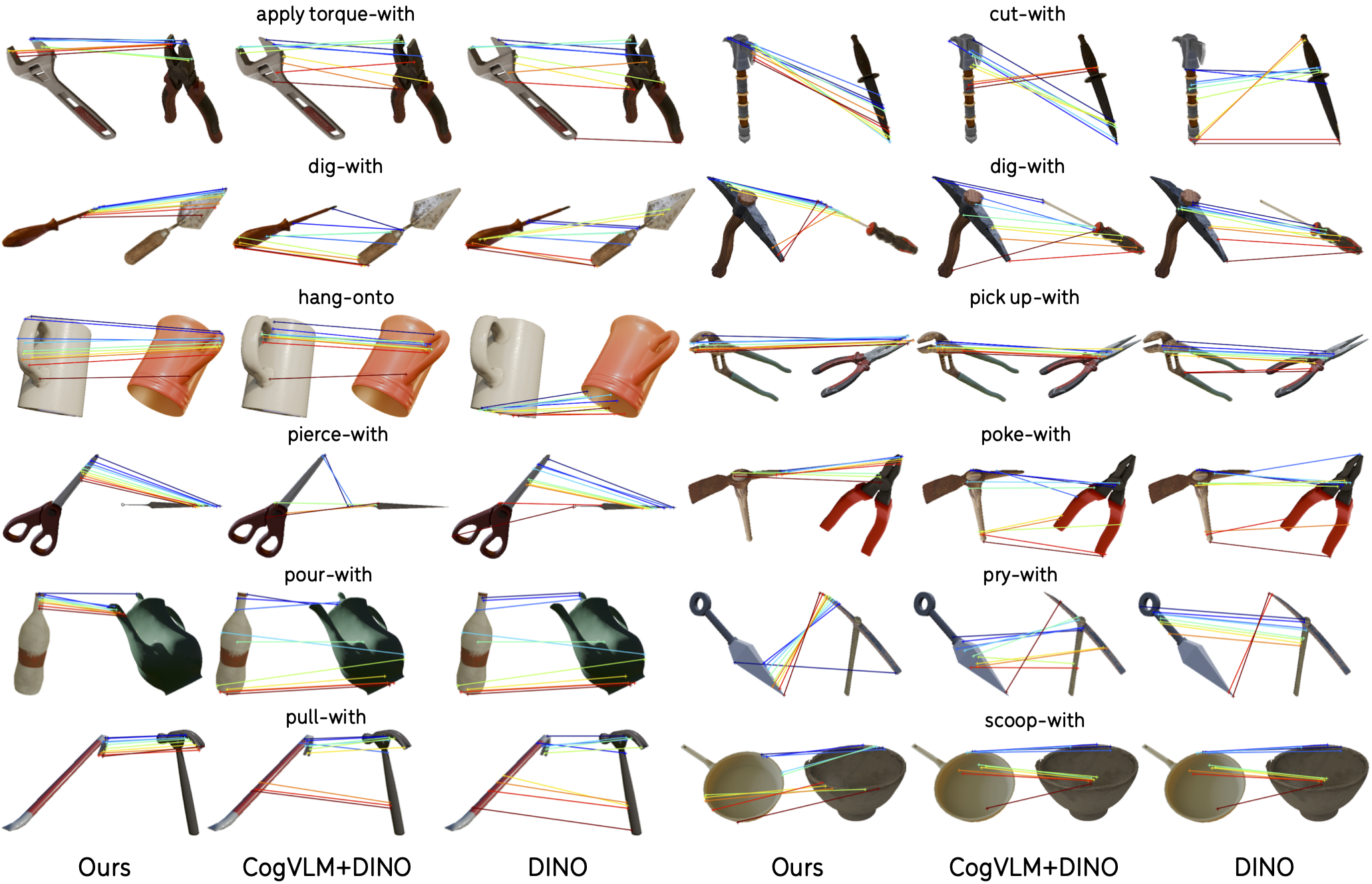}
    \vspace{-2.0em}
    \caption{\textbf{Additional Correspondence Discovery Results on Objaverse Evaluation Dataset.} We show more qualitative examples of correspondence discovery on the synthetic Objaverse evaluation dataset, comparing our model against baselines.}
    \label{fig:addnl_quali_objaverse}
    \vspace{-4.0em}
\end{figure*}

%% file: figures_tex/supp_quali_handal.tex
\begin{figure*}[t]
    \centering
    \includegraphics[width=\linewidth]{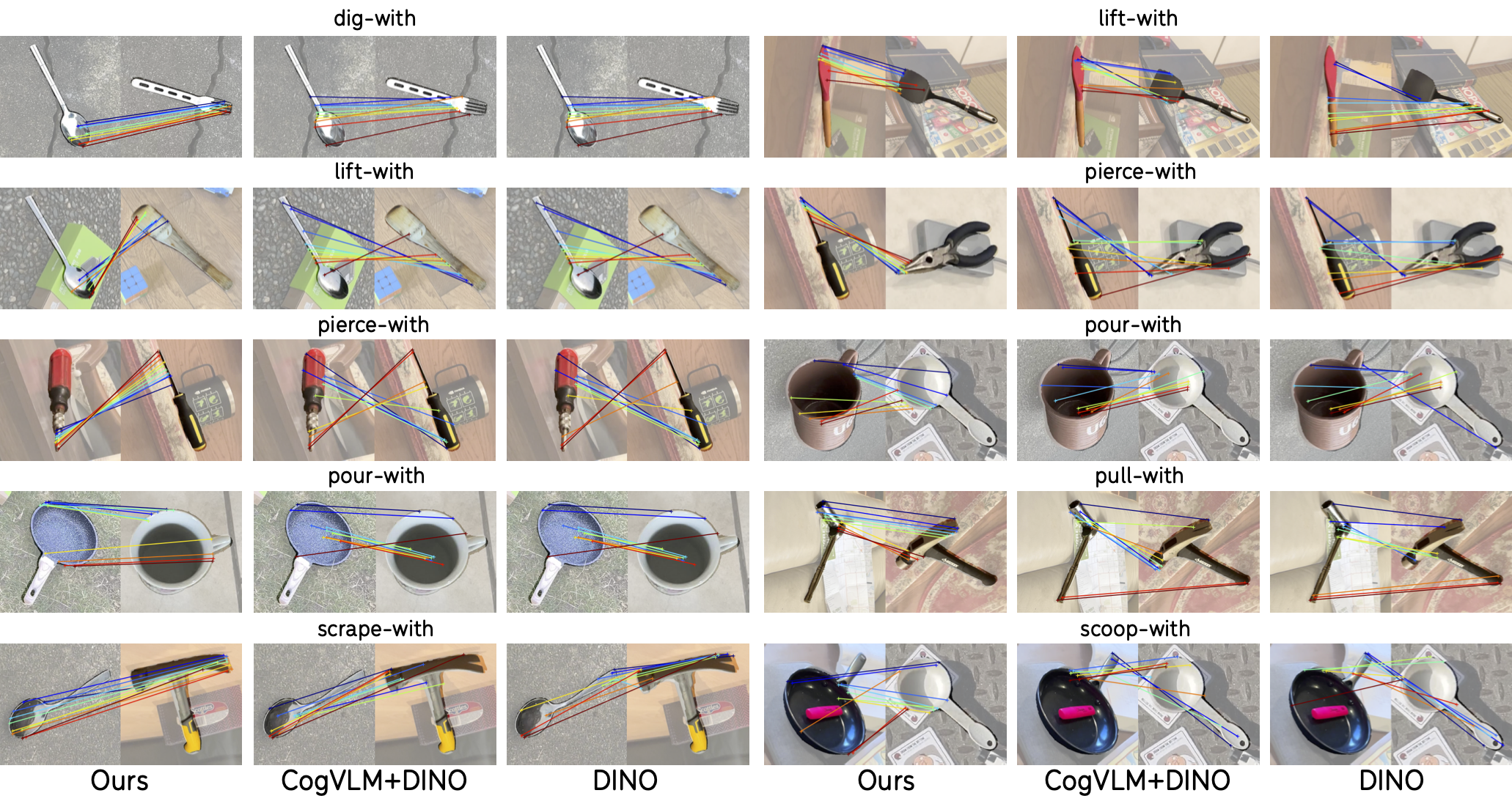}
    \vspace{-2.0em}
    \caption{\textbf{Additional Correspondence Discovery Results on HANDAL Evaluation Dataset.}}
    \label{fig:addnl_quali_handal}
    \vspace{-4.0em}
\end{figure*}

%% file: figures_tex/object_category_prompting.tex
\begin{figure}[t]
    \centering
    \includegraphics[width=\linewidth]{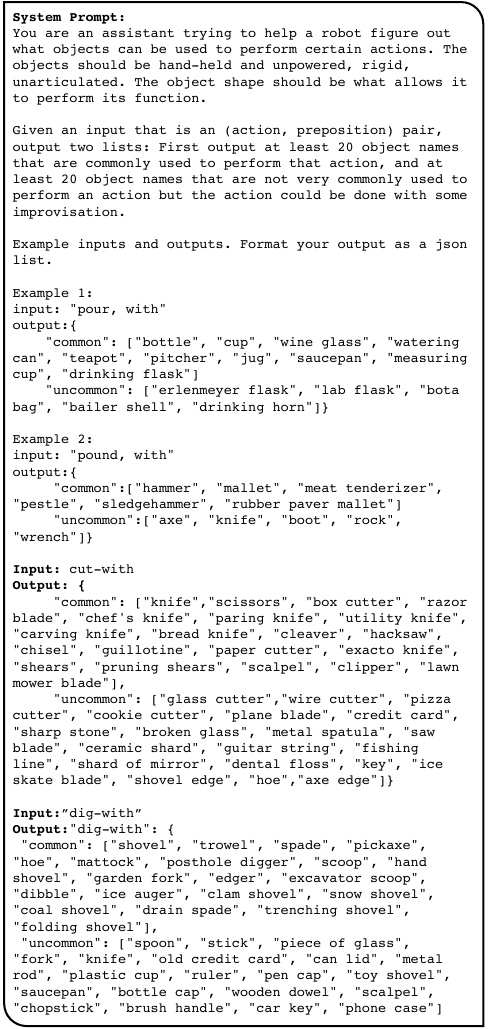}
    \caption{Our GPT-4~\cite{gpt4} prompting procedure for creating object categories given a list of functions.}
    \label{fig:object_category_prompting}
\end{figure}

%% file: figures_tex/part_desc_gen.tex
\begin{figure}[t]
    \centering
    \includegraphics[width=\linewidth]{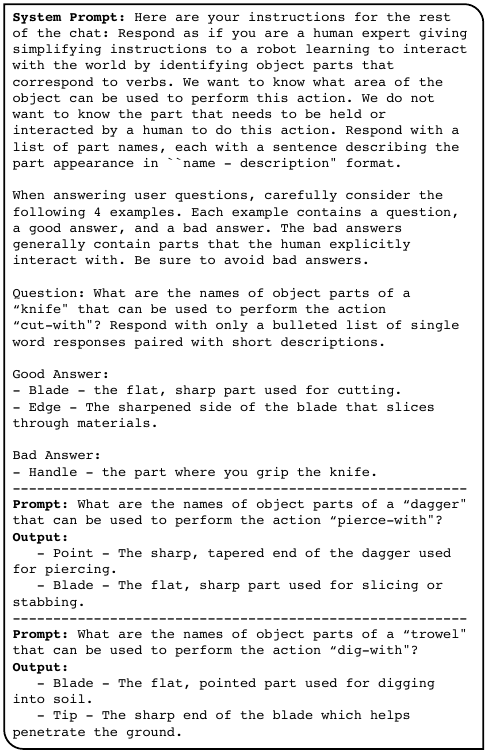}
    \vspace{-1.75em}
    \caption{Overview of our strategy for prompting GPT-4~\cite{gpt4} to obtain functional part names.}
    \label{fig:part_desc_gen}
    \vspace{-4.0em}
\end{figure}

%% file: figures_tex/llama_summarization.tex
\begin{figure}[t]
    \centering
    \includegraphics[width=\linewidth]{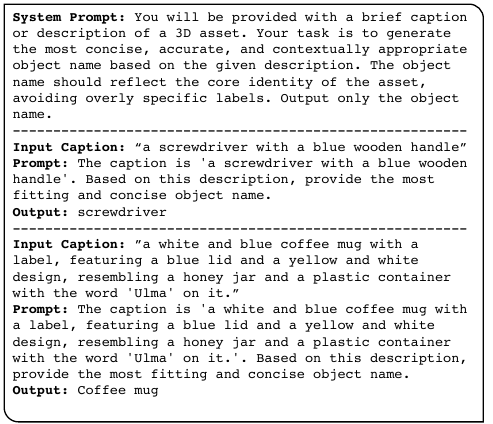}
    \vspace{-1.75em}
    \caption{Summarizing Caption3D~\cite{luo2024caption3d} captions into nouns with Llama3~\cite{dubey2024llama}. The LLM is capable of finding the noun that is the main subject of the caption.}
    \label{fig:llama_summarization}
    \vspace{-3.5em}
\end{figure}

%% file: table_tex/taxonomy.tex
\begin{table*}[ht]
\centering
\small % You can adjust the font size if necessary
\begin{tabularx}{\linewidth}{lX}
\hline
\textbf{Function} & \textbf{Objects} \\
\hline
scrape with & knife, screw, card, dagger, pen, coin, pencil, screwdriver, shovel, key, spoon, needle, scissors, pickaxe, fork, spatula, CD, hook, ruler, credit card, pitchfork, lid, pin, comb, awl, cleaver, trowel, razor blade, nail, toothpick, hockey stick, machete, rake, paddle, paper clip, license plate, hoe, corkscrew, box cutter, chisel, brush, grater, stylus, scalpel, file, letter opener, squeegee, peeler \\
press with & smartphone, bottle, shoe, stone, bowl, mug, water bottle, jug, teapot, hammer, bucket, cup, jar, book, plate, candle holder, tray, brick, pot, coffee pot, boot, flask, spoon, cutting board, pan, mallet, spatula, glass, kettle, plank, tablet, credit card, can, lid, ladle, CD case, saucepan, stamp, clipboard, paddle, pestle, hoe, meat tenderizer \\
pound with & axe, bottle, shoe, bowl, water bottle, hammer, jar, pipe, candle holder, flashlight, wrench, brick, pot, baseball bat, screwdriver, dumbbell, shovel, boot, remote control, spoon, pan, mallet, pickaxe, spatula, bowling pin, log, crowbar, can, ladle, rolling pin, gavel, cleaver, hockey stick, baton, saucepan, cricket bat, club, hairbrush, pestle, meat tenderizer \\
pierce with & screw, knife, sword, dagger, pen, pencil, drill, screwdriver, needle, scissors, pickaxe, stilettos, fork, hook, pitchfork, pin, spear, fish hook, dart, awl, chopsticks, harpoon, nail, toothpick, machete, skewer, golf tee, corkscrew, box cutter, chisel, stylus, scalpel, safety pin, letter opener \\
poke with & pipe, pencil, stick, pliers, screwdriver, key, rod, spoon, needle, pickaxe, fork, toothbrush, branch, pin, paintbrush, awl, chopsticks, nail, coat hanger, dowel, baton, antenna, toothpick, skewer, crayon, matchstick, tweezers, tongs, drumstick, stylus, stirrer, letter opener \\
mix with & knife, pen, pencil, screwdriver, rod, spoon, fork, spatula, toothbrush, branch, ruler, ladle, awl, chopsticks, baton, straw, marker, skewer, paddle, brush, tongs, whisk, scalpel, stylus, stirrer, letter opener \\
pour with & mug, bottle, shoe, bowl, jug, water bottle, teapot, bucket, cup, jar, hat, pot, coffee pot, oil can, flask, pan, watering can, hard hat, kettle, glass, can, ladle, saucepan, decanter, coconut shell \\
cut with & axe, knife, sword, dagger, key, scissors, spatula, CD, ruler, credit card, saw, cleaver, razor blade, machete, box cutter, ice skate, chisel, scalpel, pizza cutter, letter opener \\
scoop with & mug, shoe, bowl, jug, seashell, bucket, cup, hat, pot, shovel, flask, spoon, pan, hard hat, glass, ladle, trowel, saucepan, dustpan, coconut shell \\
roll onto & cylinder, mug, bottle, water bottle, cup, jar, pen, pipe, flashlight, glass, bowling pin, log, can, rolling pin, battery, lipstick, dowel, marker, roller \\
dig with & knife, stick, screwdriver, shovel, key, spoon, pickaxe, fork, ruler, awl, trowel, chopsticks, nail, paddle, hoe, dustpan, chisel, plow \\
sweep with & card, shovel, fork, spatula, broom, credit card, pitchfork, trowel, hockey stick, feather, rake, paddle, hairbrush, hoe, mop, brush, squeegee \\
pry with & knife, dagger, wrench, screwdriver, shovel, key, spoon, pickaxe, fork, spatula, ruler, crowbar, chopsticks, can opener, bottle opener, corkscrew, chisel \\
lift with & knife, seashell, plate, tray, shovel, cutting board, spoon, fork, spatula, ruler, lid, cleaver, trowel, paddle, clipboard, dustpan \\
pull with & hammer, L-bracket, hook, crowbar, fish hook, coat hanger, harpoon, carabiner, hoe, grappling hook, grabber \\
spread with & knife, card, spoon, spatula, ruler, credit card, cleaver, trowel, brush \\
brush with & broom, toothbrush, paintbrush, feather, hairbrush, mop, brush \\
write with & pen, pencil, paintbrush, lipstick, marker, crayon, stylus \\
hang onto & mug, curtain ring, hook, fish hook, coat hanger, carabiner, paper clip \\
peel with & knife, dagger, box cutter, chisel, scalpel, peeler \\
pick up with & pliers, chopsticks, tweezers, tongs, grabber \\
wedge with & axe, wedge, spatula, chisel \\
apply torque with & wrench, pliers, grabber \\
sift with & basket, strainer, colander \\
\hline
\end{tabularx}
\caption{The list of functions and the associated objects in our dataset's function/object taxonomy.}
\label{tab:functions_list}
\end{table*}

%% file: table_tex/object_counts.tex
\begin{table*}[ht]
\centering
\small % Adjust font size as necessary
\setlength{\tabcolsep}{5pt} % Adjust column separation as needed
\begin{tabularx}{\textwidth}{ >{\raggedright\arraybackslash}X r | >{\raggedright\arraybackslash}X r | >{\raggedright\arraybackslash}X r | >{\raggedright\arraybackslash}X r | >{\raggedright\arraybackslash}X r }
axe & 213 & pencil & 91 & broom & 26 & razor blade & 12 & paper clip & 8 \\
knife & 200 & brick & 85 & CD & 25 & nail & 12 & license plate & 8 \\
screw & 200 & pot & 80 & glass & 24 & chopsticks & 12 & hoe & 8 \\
card & 200 & stick & 78 & kettle & 24 & harpoon & 12 & corkscrew & 8 \\
smartphone & 200 & drill & 73 & bowling pin & 24 & coat hanger & 12 & box cutter & 8 \\
bottle & 200 & pliers & 71 & toothbrush & 24 & dowel & 12 & pestle & 8 \\
shoe & 200 & baseball bat & 66 & plank & 23 & lipstick & 12 & matchstick & 8 \\
stone & 200 & screwdriver & 56 & hook & 22 & toothpick & 11 & tweezers & 8 \\
bowl & 200 & dumbbell & 54 & tablet & 22 & hockey stick & 11 & ice skate & 8 \\
mug & 200 & coffee pot & 53 & log & 22 & machete & 11 & dustpan & 8 \\
water bottle & 200 & shovel & 51 & branch & 22 & saucepan & 11 & mop & 8 \\
jug & 200 & L-bracket & 51 & ruler & 21 & baton & 11 & colander & 8 \\
cylinder & 200 & oil can & 50 & credit card & 20 & antenna & 11 & chisel & 7 \\
teapot & 199 & key & 49 & crowbar & 19 & carabiner & 11 & brush & 7 \\
hammer & 199 & wedge & 48 & pitchfork & 18 & stamp & 10 & tongs & 7 \\
sword & 199 & boot & 47 & can & 18 & cricket bat & 10 & grater & 6 \\
seashell & 192 & remote control & 47 & lid & 17 & skewer & 10 & stylus & 6 \\
bucket & 187 & flask & 41 & ladle & 16 & golf tee & 10 & scalpel & 6 \\
dagger & 183 & rod & 41 & rolling pin & 16 & crayon & 10 & drumstick & 6 \\
cup & 181 & spoon & 39 & saw & 16 & straw & 10 & whisk & 6 \\
jar & 175 & cutting board & 39 & pin & 15 & marker & 10 & grappling hook & 6 \\
book & 173 & pan & 35 & CD case & 15 & roller & 10 & safety pin & 5 \\
basket & 163 & watering can & 34 & spear & 15 & feather & 10 & grabber & 5 \\
pen & 160 & needle & 33 & gavel & 14 & strainer & 10 & file & 4 \\
plate & 158 & scissors & 33 & fish hook & 14 & rake & 9 & stirrer & 4 \\
coin & 140 & hard hat & 33 & battery & 14 & paddle & 9 & pizza cutter & 4 \\
candle holder & 126 & mallet & 32 & comb & 13 & clipboard & 9 & letter opener & 3 \\
pipe & 126 & curtain ring & 32 & awl & 13 & club & 9 & squeegee & 2 \\
hat & 118 & pickaxe & 29 & cleaver & 13 & hairbrush & 9 & peeler & 2 \\
tray & 114 & stilettos & 29 & dart & 13 & decanter & 9 & meat tenderizer & 2 \\
flashlight & 114 & fork & 27 & paintbrush & 13 & can opener & 9 & coconut shell & 2 \\
wrench & 112 & spatula & 27 & trowel & 12 & bottle opener & 9 & plow & 1 \\
\end{tabularx}
\caption{Categories in our curated dataset and the number of assets in each category.}
\end{table*}